\documentclass[a4paper,10pt]{article}

\usepackage{graphicx}

\usepackage{fancyhdr}
\usepackage[margin=0.76in]{geometry}
\usepackage[font=small,format=plain,labelfont=bf,up,textfont=normal]{caption}
\usepackage{subcaption}
\usepackage[round]{natbib}
\usepackage{amsmath}
\usepackage{centernot}
\usepackage{mathtools}
\usepackage{amssymb}
\usepackage{enumitem}
\usepackage{setspace}
\usepackage{framed}
\DeclareMathOperator{\sign}{sign}
\DeclareMathOperator{\diag}{\textbf{diag}}

\newcommand{\transpose}[1]{\ensuremath{#1^{\scriptscriptstyle \top}}}
\newcommand{\norm}[1]{\left\lVert#1\right\rVert}
\title{\textbf{A Boundary Tilting Perspective\\ on the Phenomenon of Adversarial Examples}}

\date{}
\author{Thomas~Tanay, \textsl{Computer Science, UCL}\vspace{-0.1cm}\\
				{\normalsize \textit{thomas.tanay.13@ucl.ac.uk}}\vspace{0.1cm}\\
        Lewis~Griffin, \textsl{Computer Science, UCL}}

\begin{document}

\setlength{\abovedisplayskip}{5pt}
\setlength{\belowdisplayskip}{5pt}

\maketitle

\vspace{0.6cm}

\begin{abstract}
Deep neural networks have been shown to suffer from a surprising weakness: their classification outputs can be changed by small, non-random perturbations of their inputs. This \emph{adversarial example phenomenon} has been explained as originating from deep networks being ``too linear'' \citep{goodfellow2014explaining}. We show here that the linear explanation of adversarial examples presents a number of limitations: the formal argument is not convincing; linear classifiers do not always suffer from the phenomenon, and when they do their adversarial examples are different from the ones affecting deep networks.

We propose a new perspective on the phenomenon. We argue that adversarial examples exist when the classification boundary lies close to the submanifold of sampled data, and present a mathematical analysis of this new perspective in the linear case. We define the notion of \emph{adversarial strength} and show that it can be reduced to the \emph{deviation angle} between the classifier considered and the nearest centroid classifier. Then, we show that the adversarial strength can be made arbitrarily high independently of the classification performance due to a mechanism that we call \emph{boundary tilting}. This result leads us to defining a new taxonomy of adversarial examples. Finally, we show that the adversarial strength observed in practice is directly dependent on the level of regularisation used and the strongest adversarial examples, symptomatic of overfitting, can be avoided by using a proper level of regularisation.
\end{abstract}

\vspace{0.6cm}

\section{Introduction}

\begin{figure}[ht]
        \centering
        \begin{subfigure}[b]{0.45\textwidth}
                \includegraphics[width=\textwidth]{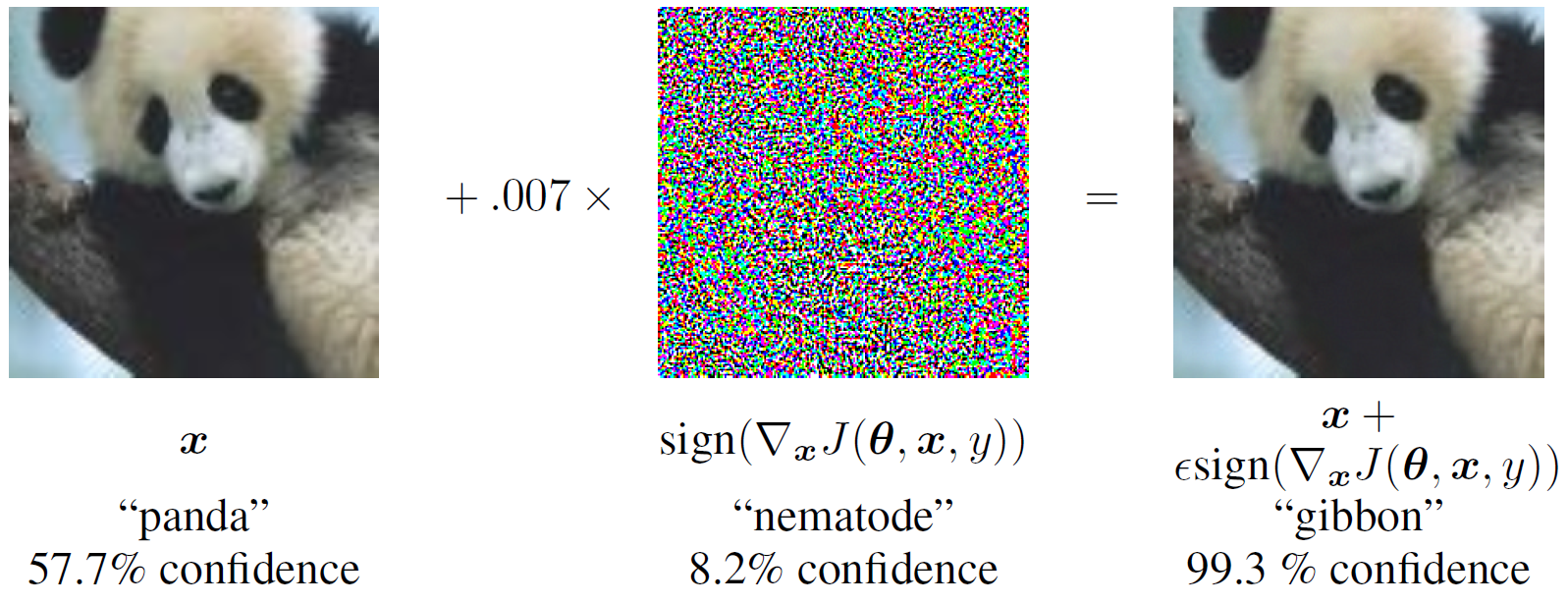}
                \caption{Adversarial example with GoogLeNet on ImageNet.}
                \label{panda}
        \end{subfigure}
	\quad\quad\quad
        \begin{subfigure}[b]{0.45\textwidth}
                \includegraphics[width=\textwidth]{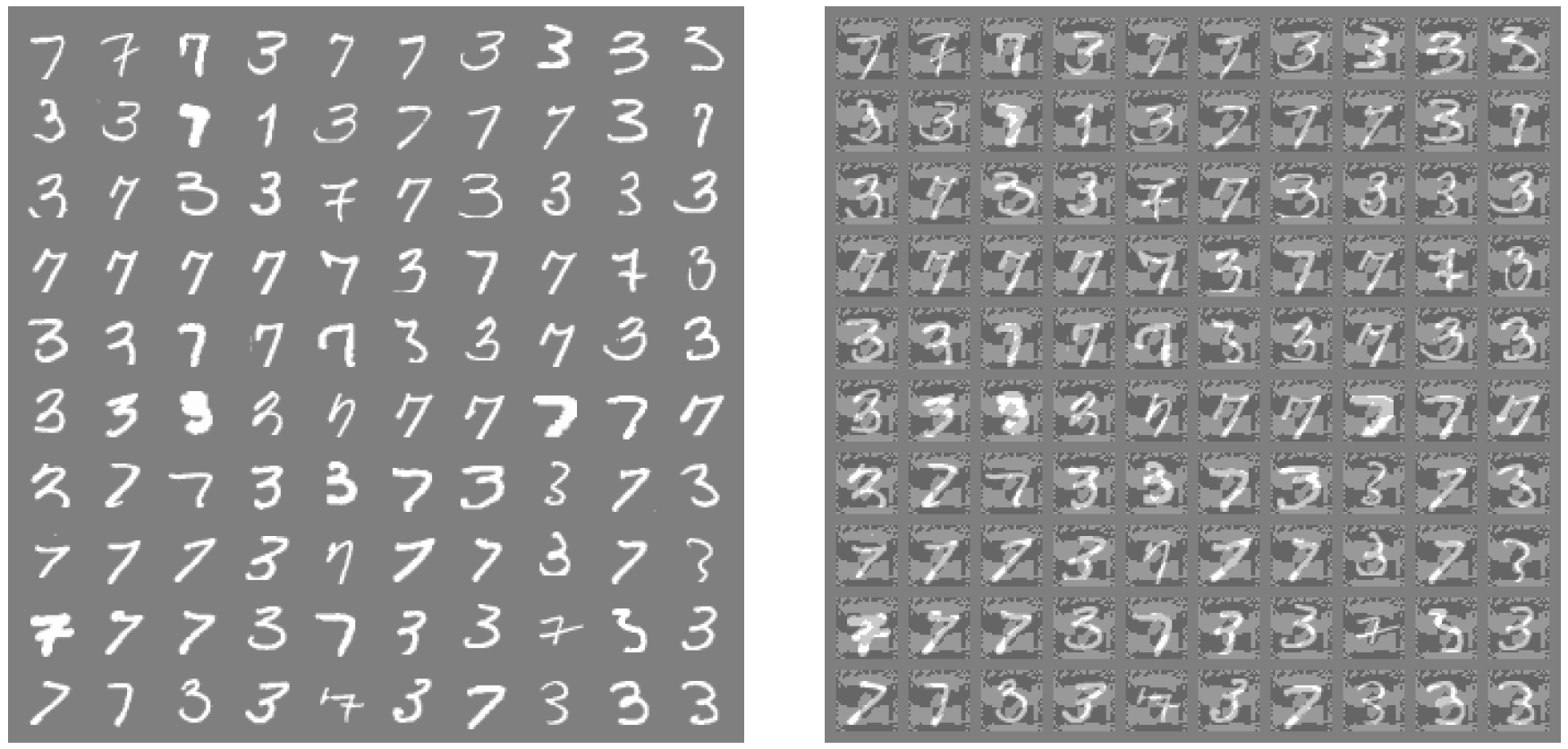}
                \caption{Left: original images from MNIST. Right: adversarial examples with logistic regression.}
                \label{mnist}
        \end{subfigure}
        \caption{Adversarial examples for two different models (from \citep{goodfellow2014explaining}).}
	\label{exampleAE}
\end{figure}

Tremendous progress has been made in the field of Deep Learning in recent years. Convolutional Neural Networks in particular, started to deliver promising results in 2012 on the ImageNet Large Scale Visual Recognition Challenge \citep{krizhevsky2012imagenet}. Since then, improvements have come at a very high pace: the range of applications has widened \citep{xu2015show,mnih2015human}, network architectures have become deeper and more complex \citep{szegedy2015going, simonyan2014very}, training methods have improved \citep{he2015deep}, and other important tricks have helped increase classification performance and reduce training time \citep{srivastava2014dropout, ioffe2015batch}. As a consequence, deep networks that are able to outperform humans are now being produced: for instance on the challenging imageNet dataset \citep{he2015delving}, or on face recognition \citep{schroff2015facenet}. Yet the same networks present a surprising weakness: their classifications are extremely sensitive to some small, non-random perturbations \citep{szegedy2013intriguing}. As a result, any correctly classified image possesses \emph{adversarial examples}: perturbed images that appear identical (or nearly identical) to the original image according to human observers --- and hence that should belong to the same class --- that are classified differently by the networks (see figure~\ref{exampleAE}). There seems to be a fundamental contradiction in the existence of adversarial examples in state-of-the-art neural networks. On the one hand, these classifiers learn powerful representations on their inputs, resulting in high performance classification. On the other hand, every image of each class is only a small perturbation away from an image of a different class. Stated differently, the classes defined in image space seem to be both well-separated and intersecting everywhere. In the following, we refer to this apparent contradiction as the \emph{adversarial examples paradox}.

In section~\ref{sec: Previous Explanations}, we present two existing answers to this paradox including the currently accepted linear explanation of \cite{goodfellow2014explaining}. In section~\ref{sec: Limitations with the Linear Explanation}, we argue that the linear explanation presents a number of limitations: the formal argument is unconvincing; we can define classes of images on which linear models do not suffer from the phenomenon; and the adversarial examples affecting logistic regression on the 3s vs 7s MNIST problem  appear qualitatively very different from the ones affecting GoogLeNet on ImageNet. In section~\ref{sec: The Boundary Tilting Perspective}, we introduce the boundary tilting perspective. We start by presenting a new pictorial solution to the adversarial examples paradox: a submanifold of sampled data, intersected by a class boundary that lies close to it, suffers from adversarial examples. Then we develop a mathematical analysis of the new perspective in the linear case. We define a strict condition for the non-existence of adversarial examples, from which we deduce a measure of \emph{strength} for the adversarial examples affecting a class of images. Then we show that the adversarial strength can be reduced to a simple parameter: the \emph{deviation angle} between the weight vector of the classifier considered and the weight vector of the nearest centroid classifier. We also show that the adversarial strength can become arbitrarily high without affecting performance when the classification boundary tilts along a component of low variance in the data. This result leads us to defining a new taxonomy of adversarial examples. Finally, we show experimentally using SVM that the adversarial strength observed in practice is controlled by the level of regularisation used. With very high regularisation, the phenomenon of adversarial examples is minimised and the classifier defined converges towards the nearest centroid classifier. With very low regularisation however, the training data is overfitted by boundary tilting, leading to the existence of strong adversarial examples.

\section{Previous Explanations}
\label{sec: Previous Explanations}

\subsection{Low-probability ``pockets'' in the manifold}
 
In \citep{szegedy2013intriguing}, the existence of adversarial examples was regarded as an intriguing phenomenon. No detailed explanation was proposed, and only a simple analogy was introduced: 
\begin{quote}
\textit{``Possible explanation is that the set of adversarial negatives is of extremely low probability, and thus is never (or rarely) observed in the test set, \emph{yet it is dense (much like the rational numbers)}, and so it is found virtually near every test case''} [emphasis added]
\end{quote}

Using the mathematical concept of density, and the example of the rational numbers in particular, we can indeed define a classifier that suffers from the phenomenon of adversarial examples. Consider the classifier $\mathcal{C}$ operating on the real numbers with the following decision rule for a test number $x$:
\begin{itemize}[parsep=0cm, itemsep=0cm, topsep=0cm]
\item $x$ belongs to \raisebox{0.2ex}{$+$} if it is positive irrational or negative rational.
\item $x$ belongs to \raisebox{0.2ex}{$-$} if it is negative irrational or positive rational.
\end{itemize}
On a test set selected at random among real numbers, $\mathcal{C}$ discriminates perfectly between positive and negative numbers: real numbers contain infinitely more irrational numbers than rational numbers and for whatever test number $x$ we choose at random among real numbers, $x$ is infinitely likely to be irrational, and thus correctly classified. Yet $\mathcal{C}$ suffers from the phenomenon of adversarial examples: since the set of rational numbers is dense in the set of real numbers, $x$ is infinitely close to rational numbers that constitute adversarial examples.

\paragraph{}
The rational numbers analogy is interesting, but it leaves one important question open: why would deep networks define decision rules that are in any way as strange as the one defined by our example classifier $\mathcal{C}$? By what mechanism should the low-probability ``pockets'' be created? Without attempting to provide a detailed answer, \cite{szegedy2013intriguing} suggested that it was made possible by the high non-linearity of deep networks.

\subsection{Linear explanation}

\cite{goodfellow2014explaining} subsequently provided a more detailed analysis of the phenomenon, and introduced the linear explanation --- currently generally accepted. Their explanation relies on a new analogy:
\begin{quote}
\textit{``We can think of this as a sort of \emph{`accidental steganography'}, where a linear model is forced to attend exclusively to the signal that aligns most closely with its weights, even if multiple signals are present and other signals have much greater amplitude.''} [emphasis added]
\end{quote}

Given an input $x$ and an adversarial example $\boldsymbol{\tilde{x}} = \boldsymbol{x} + \boldsymbol{\eta}$ where $\boldsymbol{\eta}$ is subject to the constraint $\|\boldsymbol{\eta}\|_\infty < \epsilon$, the argument is the following:
\begin{quote}
\textit{``Consider the dot product between a weight vector $\boldsymbol{w}$ and an adversarial example $\boldsymbol{\tilde{x}}$:
$$\transpose{\boldsymbol{w}} \cdot \boldsymbol{\tilde{x}} = \transpose{\boldsymbol{w}} \cdot \boldsymbol{x} + \transpose{\boldsymbol{w}} \cdot \boldsymbol{\eta}$$
The adversarial perturbation causes the activation to grow by $\transpose{\boldsymbol{w}} \cdot \boldsymbol{\eta}$. We can maximise this increase subject to the max norm constraint on $\boldsymbol{\eta}$ by assigning $\boldsymbol{\eta} = \epsilon\,\boldsymbol{sign(w)}$. If $\boldsymbol{w}$ has $n$ dimensions and the average magnitude of an element of the weight vector is $m$, then the activation will grow by $\epsilon\,m\,n$. Since $\|\boldsymbol{\eta}\|_\infty$ does not grow with the dimensionality of the problem but the change in activation caused by the perturbation by $\boldsymbol{\eta}$ can grow linearly with $n$, then for high dimensional problems, we can make many infinitesimal changes to the input that add up to one large change to the output.''}
\end{quote}

The authors concluded that ``a simple linear model can have adversarial examples if its input has sufficient dimensionality''. This argument was followed with the observation that small linear movements in the direction of the sign of the gradient (with respect to the input image) can cause deep networks to change their predictions, and hence that ``linear behaviour in high-dimensional spaces is sufficient to cause adversarial examples''.

\begin{framed}
\noindent \textbf{Technical remarks:}
\begin{enumerate}[parsep=0.1cm, itemsep=0.1cm, topsep=0.1cm]
\item What norm should be used to evaluate the magnitude of a small perturbation? The image perturbations used to generate adversarial examples are typically measured with a norm that does not necessarily match perceptual magnitude. For instance, \cite{goodfellow2014explaining} use the infinity norm, based on the idea that digital measuring devices are insensitive to small perturbations whose infinity norm is below a certain threshold (because of digital quantization). This is a reasonable but arbitrary choice. We might consider other norms more adapted (such as 1- or 2-norm) --- because for human observers, the magnitude of a perturbation does not only depend on the maximum change along individual pixels but also on the number of changing pixels\footnotemark. This is a fairly technical point of little importance in practice, except for determining the specific direction in which to move when looking for adversarial examples. We use the 2-norm, so that the direction we move in is simply the direction of the gradient. In other words, we create adversarial examples by adding the quantity $\epsilon\,\boldsymbol{w}/\|\boldsymbol{w}\|_2$ to the input image, instead of adding the quantity $\epsilon\,\boldsymbol{sign(w)}$, as one does for the infinity norm.
\item In previous works, the phenomenon of adversarial examples in linear classification was investigated using logistic regression \citep{szegedy2013intriguing, goodfellow2014explaining}. In the present study, we use another standard linear classifier: support vector machine (SVM) with linear kernel. The two methods are largely equivalent but we prefer SVM for its geometrical interpretation, more adapted to the boundary tilting perspective we introduce in the following.
\end{enumerate}
\end{framed}
\footnotetext{A perturbation of $\epsilon$ on the pixel in the top left corner of an image does not have the same perceptual magnitude as a perturbation of $\epsilon$ across the entire image. Yet the infinity norm gives the same magnitude to the two perturbations.}

\section{Limitations with the Linear Explanation}
\label{sec: Limitations with the Linear Explanation}

\subsection{An unconvincing argument}
\label{sec: An unconvincing argument}

The idea of accidental steganography is a seducing intuition that seems to illustrate well the phenomenon of adversarial examples. Yet the argument is unconvincing: small perturbations do not provoke changes in activation that grow linearly with the dimensionality of the problem, \emph{when they are considered relatively to the activations themselves}. Consider the dot product between a weight vector $\boldsymbol{w}$ and an adversarial example $\boldsymbol{\tilde{x}}$ again: $\transpose{\boldsymbol{w}} \cdot \boldsymbol{\tilde{x}} = \transpose{\boldsymbol{w}} \cdot \boldsymbol{x} + \transpose{\boldsymbol{w}} \cdot \boldsymbol{\eta}$. As we have seen before, the change in activation $\transpose{\boldsymbol{w}} \cdot \boldsymbol{\eta}$ grows linearly with the problem; but so does the activation $\transpose{\boldsymbol{w}} \cdot \boldsymbol{x}$ (provided that the weight and pixel distributions in $\boldsymbol{w}$ and $\boldsymbol{x}$ stay unchanged), and the ratio between the two quantities stays constant.

\begin{figure}[ht]
        \centering
        \begin{subfigure}[b]{0.49\textwidth}
                \includegraphics[width=\textwidth]{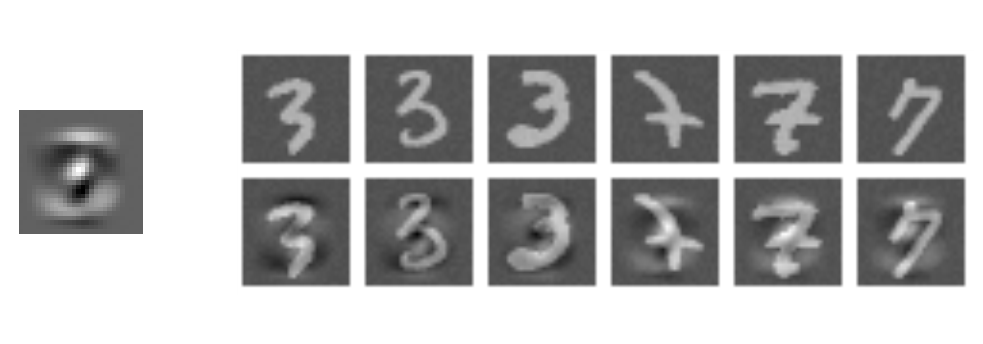}
                \caption{3s vs 7s MNIST problem with an image size of ${28 \times 28}$. Left: weight vector defined by linear SVM. Right: example digits (top) and their adversarial examples (bottom).}
                \label{AEs28}
        \end{subfigure}
	\enspace
        \begin{subfigure}[b]{0.49\textwidth}
                \includegraphics[width=\textwidth]{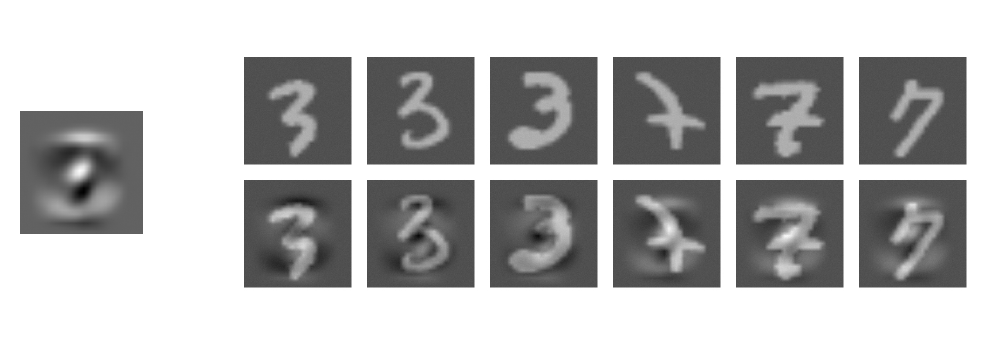}
                \caption{3s vs 7s MNIST problem with an image size of $200 \times 200$. Left: weight vector defined by linear SVM. Right: the same example digits (top) and their adversarial examples (bottom).}
                \label{AEs200}
        \end{subfigure}
        \caption[Caption for LOF]{Increasing the dimensionality of the problem does not make the phenomenon of adversarial examples worse. Whether the image size is $28 \times 28$ or $200 \times 200$, the weight vector found by linear SVM looks very similar to the one found by logistic regression in \citep{goodfellow2014explaining}. The two SVM models have an error rate of $2.7\%$\footnotemark. The magnitude $\epsilon$ of the perturbations has been chosen in both cases such that $99\%$ of the digits in the test set are misclassified ($\epsilon_{28} = 4.6, \epsilon_{200} = 31. \approx \epsilon_{28} \times 200/28$)}
	\label{MNISTdim}
\end{figure}

\paragraph{}
We illustrate this by performing linear classification on a modified version of the 3s vs 7s MNIST problem where the image size has been increased to $200 \times 200$. We generated the new dimensions by linear interpolation and increased variability by adding some noise to the original and the modified datasets (random perturbations between $[-0.05,0.05]$ on every pixel). The results for the two image sizes look strikingly similar (see figure~\ref{MNISTdim}). Importantly, increasing the image resolution has no influence on the perceptual magnitude of the adversarial perturbations, even if the dimension of the problem has been multiplied by more than 50.

\footnotetext{Better error rates can be obtained by using less regularisation, as shown in section~\ref{sec: Return to MNIST}.}

In sum, the dimensionality argument does not hold: high dimensional problems are not necessarily more prone to the phenomenon of adversarial examples. Without this central result however, can we still maintain that linear behaviour is sufficient to cause adversarial examples?

\subsection{Linear behaviour is not sufficient to cause adversarial examples}
\label{sec: Linear behaviour is not sufficient to cause adversarial examples}

\begin{figure}[ht]
  \centering
  \includegraphics[width=\textwidth]{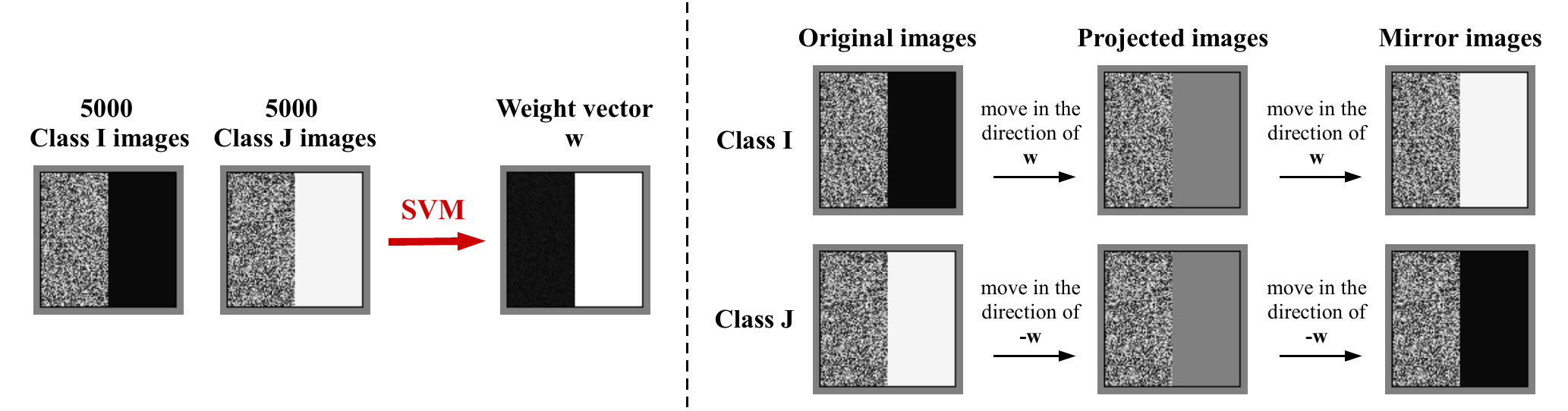}
  \caption{Toy problem of two classes $I$ and $J$ that do not suffer from the phenomenon of adversarial examples. When we follow the procedure that normally leads to the creation of adversarial examples, we get instead real instances of images that belong to the other class. We call the images on the boundary the \emph{projected images} and the images with opposed classification score the \emph{mirror images}.}
  \label{toyExample}
\end{figure}

According to the linear explanation of \cite{goodfellow2014explaining}, linear behaviour itself is responsible for the existence of adversarial examples. If we take this explanation literally, then  we expect all linear classification problems to suffer from the phenomenon. Yet we can find classes of images for which adversarial examples do not exist at all. Consider the following toy problem (figure~\ref{toyExample}).

\bigskip
\noindent Let $I$ and $J$ be two classes of images of size $100\times100$ defined as follow:
\begin{description}[parsep=0.1cm, itemsep=0.1cm, topsep=0.2cm]
\item[Class $I$.] Left half-image noisy (random pixel values in $[0, 1]$) and right half-image black (pixel value: 0).
\item[Class $J$.] Left half-image noisy (random pixel values in $[0, 1]$) and right half-image white (pixel value: 1).
\end{description}

\medskip
\noindent If we train a linear SVM on $5000$ images of each class, we achieve perfect separation of the training data with full generalisation to novel test data. When we look at the weight vector $\boldsymbol{w}$ defined by SVM, we notice that it correctly represents the feature separating the two classes: it ignores the left half-image (all weights near zero) and takes into consideration the entire right half-image (all weights near 1). As a result, adversarial examples do not exist. Indeed, if we take an image in one of the two classes and move in the gradient direction until we reach the class boundary, then we get an image that is also perceived as being between the two classes according to human observers (grey right half-image); and if we continue to move in the gradient direction until we reach a confidence level that the new image belongs to the new class equal to the confidence level that the original image belonged to the original class, then we get an image that is also perceived as belonging to the new class according to human observers.

This toy problem is very artificial and the point we make from it might seem little convincing for the moment, but it should not be disputed that there is a priori nothing in the current linear explanation that allows us to predict which classes of images will suffer from the phenomenon of adversarial examples, and which will not. In the following section we consider a more realistic problem: MNIST. We will return to the toy problem in section~\ref{sec: Return to the toy problem}.

\subsection{Linear classification on MNIST. Are these examples really adversarial?}
\label{sec: Linear classification on MNIST. Are these examples really adversarial?}

A key argument in favour of the linear explanation of adversarial examples was that logistic regression also suffers from the phenomenon. In contrast, we argue here that what happens with linear classifiers on MNIST is very different from what happens with deep networks on ImageNet.

\paragraph{}
The first difference between the two situations is very clear: the adversarial perturbations have a much higher magnitude and are very perceptible by human observers in the case of linear classifiers on MNIST (see figure~\ref{exampleAE}). Importantly, the image resolution cannot account for this difference: increasing the size of the MNIST images does not influence the perceptual magnitude of the adversarial perturbations (as shown in section~\ref{sec: An unconvincing argument}). Not only does the linear explanation unreliably predict whether the phenomenon of adversarial examples will occur on a specific dataset (as shown in section~\ref{sec: Linear behaviour is not sufficient to cause adversarial examples}), it also cannot predict the magnitude of the adversarial perturbations necessary to make the classifier change its predictions when the phenomenon \emph{does occur}. 

\begin{figure}[ht]
        \centering
        \begin{subfigure}[t]{0.35\textwidth}
                \includegraphics[width=\textwidth]{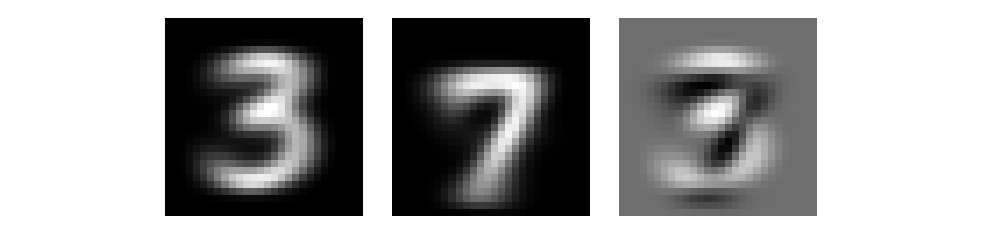}
                \caption{Average 3 (left) and average 7 (middle) on the MNIST training data. Difference between the two (right).}
                \label{mean3mean7}
        \end{subfigure}
	\quad\quad\quad
        \begin{subfigure}[t]{0.55\textwidth}
                \includegraphics[width=\textwidth]{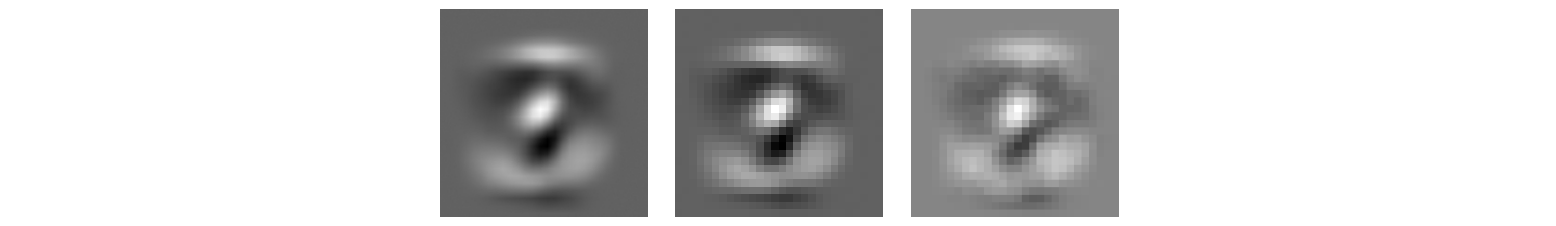}
                \caption{Weights vectors: SVM on the $200 \times 200$ images (left), SVM on the $28 \times 28$ images (middle), logistic regression in \citep{goodfellow2014explaining}, (right).}
                \label{Ws}
        \end{subfigure}
        \caption{The weight vectors found by linear models resemble the average 3 of the MNIST training data to which the average 7 has been subtracted.}
	\label{nearestCentroid}
\end{figure}

\paragraph{}
Another important difference between the adversarial examples shown in \citep{goodfellow2014explaining} for 
GoogLeNet on ImageNet and the ones shown for logistic regression on MNIST concerns the appearance of the adversarial perturbations. With GoogLeNet on ImageNet, the perturbation is dominated by high-frequency structure which cannot be meaningfully interpreted; with logistic regression on MNIST, the perturbation is low-frequency dominated and although \cite{goodfellow2014explaining} argue that it is ``not readily recognizable to a human observer as having anything to do with the relationship between 3s and 7s'', we believe that it can be meaningfully interpreted: the weight vector found by logistic regression points in a direction that is close to passing through the mean images of the two classes, thus defining a decision boundary similar to the one of a nearest centroid classifier (see figure~\ref{nearestCentroid}).

Simple linear models defined by SVM or logistic regression can be deceived on MNIST by perturbations that are visually perceptible and that look roughly like the weight vector of the nearest centroid classifier. This result is hardly surprising and does not help explain why much more sophisticated models --- such as deep networks --- can be deceived by imperceptible perturbations which look to human observers like random noise. Clearly, the linear explanation is still incomplete.

\section{The Boundary Tilting Perspective}
\label{sec: The Boundary Tilting Perspective}

\subsection{Pictorial solution to the adversarial examples paradox}

In previous sections, we rejected the linear explanation of \cite{goodfellow2014explaining}: high dimension is insufficient to explain the phenomenon of adversarial examples and linear models seem to suffer from a weaker type of adversarial examples than deep networks. Without the linear explanation however, the adversarial examples paradox persists: how can two classes of images be well separated, if every element of each class is close to an element of the other class?

\begin{figure}[ht]
        \centering
        \begin{subfigure}[b]{0.48\textwidth}
                \includegraphics[width=\textwidth]{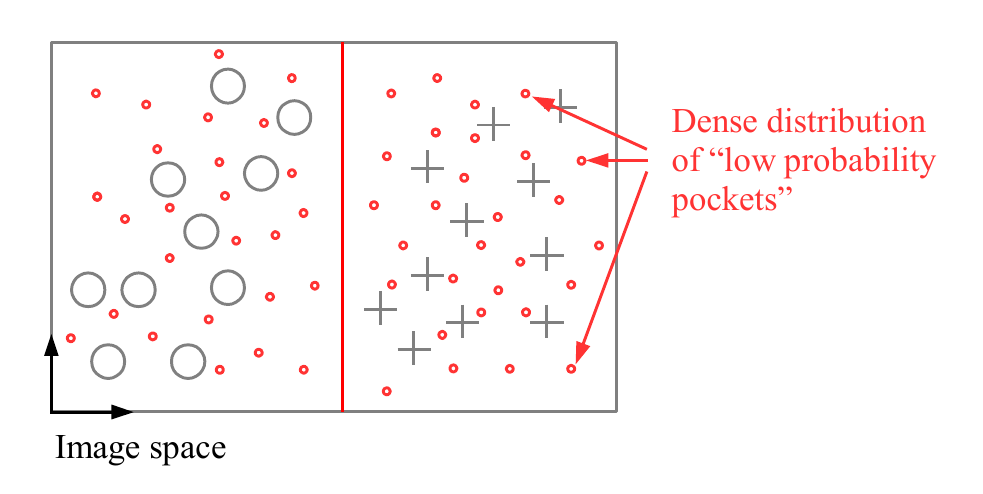}
                \caption{The solution proposed in \citep{szegedy2013intriguing}. Adversarial examples are possible because the image space is densely filled with low probability adversarial pockets.}
                \label{pocketView}
        \end{subfigure}
				\quad\enspace
        \begin{subfigure}[b]{0.48\textwidth}
                \includegraphics[width=\textwidth]{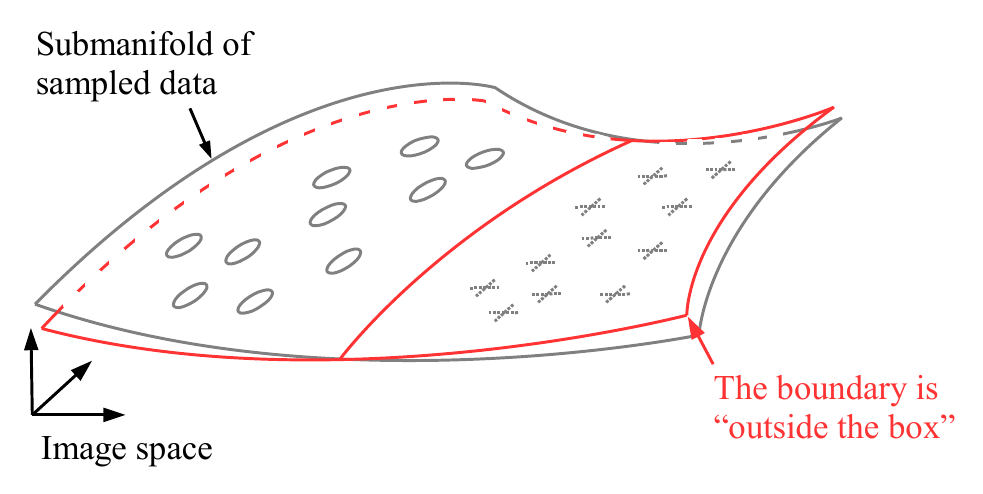}
                \caption{The solution we propose. Adversarial examples are possible because the class boundary extends beyond the submanifold of sample data and can be --- under certain circumstances --- lying close to it.}
                \label{newPerspective}
        \end{subfigure}
	\caption{Schematic representations of two solutions to the adversarial examples paradox.}
\end{figure}

\paragraph{}
In figure~\ref{pocketView}, we present a schematic representation of the solution proposed in \citep{szegedy2013intriguing}: the classes \raisebox{-0.25ex}{\LARGE $\circ$} and \raisebox{0.2ex}{$+$} are well separated, but every element of each class is very close to an element of the other class because low probability adversarial pockets are densely distributed in image space. In figure~\ref{newPerspective}, we introduce a new solution. First, we observe that the data sampled in the training and test sets only extends in a submanifold of the image space. A class boundary can intersect this submanifold such that the two classes are well separated, but will also extend beyond it. Under certain circumstances, the boundary might be lying very close to the data, such that small perturbations directed towards the boundary might cross it. 

Note that in the low dimensional representation of figure~\ref{newPerspective}, randomly perturbed images are likely to cross the class boundary. In higher dimension however, the probability that a random perturbation moves exactly in the direction of the boundary is low, such that images that are close to it (and thus sensitive to adversarial perturbations), are robust to random perturbations, in accordance with the results in \citep{szegedy2013intriguing}.

\subsection{Adversarial examples in linear classification}
\label{sec: Adversarial examples in linear classification}

The drawing of figure~\ref{newPerspective} is, of course, a severe oversimplification of the reality --- but it is a useful one. As we noticed already, it is a low dimensional impression of a phenomenon happening in much higher dimension. It also misrepresents the complexity of real data distributions and the highly non-linear nature of the class boundary defined by a state-of-the-art classifier. Yet it is useful because it allows us to make important predictions. First, the drawing is compatible with a flat class boundary and no non-linearity is required (contrary to the view relying on the presence of low probability pockets). Hence the phenomenon of adversarial examples should be observable in linear classification. At the same time, linear behaviour is not sufficient for the phenomenon to occur either: the class boundary needs to ``be tilted'' and lie close to the data. In the following, we propose a mathematical analysis of this boundary tilting explanation in linear classification. We start by giving a strict condition for the non-existence of adversarial examples, from which we deduce a measure of \emph{strength} for the adversarial examples affecting a class of images. We also show that the adversarial strength can be reduced to a simple parameter: the \emph{deviation angle} between the classifier considered and the nearest centroid classifier. Then, we introduce the \emph{boundary tilting mechanism} and show that it can lead to adversarial examples of arbitrary strength without affecting classification performance. Finally, we propose a new taxonomy of adversarial examples.

\subsubsection{Condition for the non-existence of adversarial examples}

In the standard procedure, adversarial examples are found by moving along the gradient direction by a magnitude $\epsilon$ chosen such that 99\% of the data is misclassified \citep{goodfellow2014explaining}. The smaller $\epsilon$ is, the more ``impressive'' the resulting adversarial examples. This approach is meaningful when $\epsilon$ is very small --- but as $\epsilon$ grows, when should one stop considering the images obtained as adversarial examples? When they start to actually look like images of the other class? Or when the adversarial perturbation starts to be perceptible to the human eye? Here, we introduce a strict condition for the non-existence of adversarial examples.

\paragraph{}
Let $I$ and $J$ be two classes of images, and $\mathcal{C}$ a hyperplane boundary defining a linear classifier in $\mathbb{R}^n$. $\mathcal{C}$ is formally specified by a normal weight vector $\boldsymbol{c}$ (we assume that $\|\boldsymbol{c}\|_2 =1$) and a bias $c_0$. For any image $\boldsymbol{x}$ in $\mathbb{R}^n$, we define:

\begin{itemize}[parsep=0cm, itemsep=0.1cm, topsep=0.1cm]
\item The \emph{classification score} of $\boldsymbol{x}$ through $\mathcal{C}$ as: \enspace $d(\boldsymbol{x},\mathcal{C}) = \boldsymbol{x} \cdot \boldsymbol{c} + c_0$\\
$d(\boldsymbol{x},\mathcal{C})$ is the signed distance between $\boldsymbol{x}$ and $\mathcal{C}$.\\
$\boldsymbol{x}$ is classified in $I$ if $d(\boldsymbol{x},\mathcal{C}) \leq 0$ and $\boldsymbol{x}$ is classified in $J$ if $d(\boldsymbol{x},\mathcal{C}) \geq 0$. 
\item The \emph{projected image} of $\boldsymbol{x}$ on $\mathcal{C}$ as: \enspace $\boldsymbol{p}(\boldsymbol{x},\mathcal{C}) = \boldsymbol{x} - d(\boldsymbol{x},\mathcal{C})\,\boldsymbol{c}$\\
$\boldsymbol{p}(\boldsymbol{x},\mathcal{C})$ is the nearest image $\boldsymbol{y}$ lying on $\mathcal{C}$ (i.e. such that $d(\boldsymbol{y},\mathcal{C}) = 0$).
\item The \emph{mirror image} of $\boldsymbol{x}$ through $\mathcal{C}$ as: \enspace $\boldsymbol{m}(\boldsymbol{x},\mathcal{C}) = \boldsymbol{x} - 2\,d(\boldsymbol{x},\mathcal{C})\,\boldsymbol{c}$\\
$\boldsymbol{m}(\boldsymbol{x},\mathcal{C})$ is the nearest image $\boldsymbol{y}$ with opposed classification score (i.e. such that ${d(\boldsymbol{y},\mathcal{C})=-d(\boldsymbol{x},\mathcal{C})}$).
\item The \emph{mirror class} of $I$ through $\mathcal{C}$ as: \enspace $m(I,\mathcal{C}) = \{\boldsymbol{m}(\boldsymbol{x},\mathcal{C}) \enspace | \enspace \forall \boldsymbol{x} \in I\}$
\end{itemize}

\paragraph{}
Suppose that $\mathcal{C}$ \emph{does not} suffer from adversarial examples. Then for every image $\boldsymbol{x}$ in $I$, the projected image $\boldsymbol{p}(\boldsymbol{x},\mathcal{C})$ must lie exactly between the classes $I$ and $J$. Since $\boldsymbol{p}(\boldsymbol{x},\mathcal{C})$ is the midpoint between $\boldsymbol{x}$ and the mirror image $\boldsymbol{m}(\boldsymbol{x},\mathcal{C})$, we can say that $\boldsymbol{p}(\boldsymbol{x},\mathcal{C})$ lies exactly between $I$ and $J$ iff $\boldsymbol{m}(\boldsymbol{x},\mathcal{C})$ belongs to $J$. Hence we can say that the class $I$ does not suffer from adversarial examples iff $m(I,\mathcal{C}) \subset J$. Similarly, we can say that the class $J$ does not suffer from adversarial examples iff $m(J,\mathcal{C}) \subset I$. Since the mirror operation is involutive, we have $m(I,\mathcal{C}) \subset J \Rightarrow I \subset m(J,\mathcal{C})$ and $m(J,\mathcal{C}) \subset I \Rightarrow J \subset m(I,\mathcal{C})$. Hence:
$$\boxed{\text{$\mathcal{C}$ \underline{does not} suffer from adversarial examples} \enspace \Leftrightarrow \enspace m(I,\mathcal{C}) = J \text{ and } m(J,\mathcal{C}) = I}$$
The non-existence of adversarial examples is equivalent to the classes $I$ and $J$ being mirror classes of each other through $\mathcal{C}$, or to the mirror operator $\boldsymbol{m}(\cdot,\mathcal{C})$ defining a bijection between $I$ and $J$. Conversely, we say that a classification boundary $\mathcal{C}$ suffers from adversarial examples iff ${m(I,\mathcal{C}) \neq J}$ and ${m(J,\mathcal{C}) \neq I}$. In that case, we call \emph{adversarial examples affecting $I$} the elements of $m(I,\mathcal{C})$ that are not in $J$ and we call \emph{adversarial examples affecting $J$} the elements of $m(J,\mathcal{C})$ that are not in $I$.

\newpage
\subsubsection{Strength of the adversarial examples affecting a class of images}

As discussed before, the magnitude $\epsilon$ of the adversarial perturbations used in the standard procedure is a good measure of how ``impressive'' or ``strong'' the adversarial examples are. Unfortunately, this measure is only meaningful for small values. We introduce here a measure of \emph{strength} that is valid on the entire spectrum of the adversarial example phenomenon.

\begin{description}
\item \underline{Maximum strength.} Let us note $\boldsymbol{i}$ and $\boldsymbol{j}$ the mean images of $I$ and $J$ respectively. For an element $\boldsymbol{x}$ in $I$, the ``strength'' of the adversarial example $\boldsymbol{m}(\boldsymbol{x},\mathcal{C})$ is maximised when the distance $\norm{\boldsymbol{x} - \boldsymbol{m}(\boldsymbol{x},\mathcal{C})}$ tends to 0 (this is equivalent to $\epsilon$ tending to 0 in the standard procedure). Averaging over all the elements of $I$, we can say that \emph{the strength of the adversarial examples affecting $I$ is maximised when the distance $\norm{\boldsymbol{i} - \boldsymbol{m}(\boldsymbol{i},\mathcal{C})}$ tends to 0} (see figure~\ref{strongAE}).

\begin{figure}[ht]
  \centering
  \includegraphics[width= \textwidth]{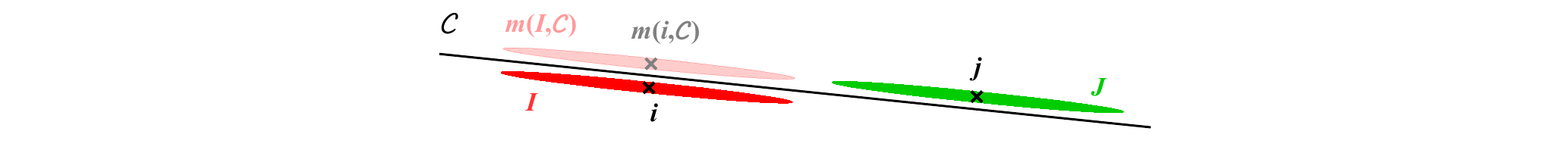}
  \caption{The smaller the distance $\norm{\boldsymbol{i} - \boldsymbol{m}(\boldsymbol{i},\mathcal{C})}$, the stronger the adversarial examples affecting $I$.}
  \label{strongAE}
\end{figure}

Remark that $\norm{\boldsymbol{i} - \boldsymbol{m}(\boldsymbol{i},\mathcal{C})} = 2\,|d(\boldsymbol{i},\mathcal{C})|$ and consider the projections of the elements in $I$ along the direction $\boldsymbol{c}$: their mean value is $d(\boldsymbol{i},\mathcal{C})$ and we note $\sigma$ their standard deviation. Consider in particular the elements $X$ in $I$ that are more than one standard deviation away from the mean in the direction $\boldsymbol{c}$: for each element $\boldsymbol{x}$ in $X$ we have $d(\boldsymbol{i},\mathcal{C}) + \sigma \leq d(\boldsymbol{x},\mathcal{C})$. If there are no strong outliers in the data, a significant proportion of the elements of $I$ belongs to $X$, and if the classifier $\mathcal{C}$ has a good performance, some of the elements in $X$ must be correctly classified in $I$, i.e. some elements in $X$ must verify $d(\boldsymbol{x},\mathcal{C}) < 0$. Hence we must have $d(\boldsymbol{i},\mathcal{C}) + \sigma < 0$ and $|d(\boldsymbol{i},\mathcal{C})| > \sigma$. We can thus write: $\norm{\boldsymbol{i} - \boldsymbol{m}(\boldsymbol{i},\mathcal{C})} = 2\,|d(\boldsymbol{i},\mathcal{C})| > 2\,\sigma$. The strength of the adversarial examples affecting $I$ is maximised ($\norm{\boldsymbol{i} - \boldsymbol{m}(\boldsymbol{i},\mathcal{C})} \to 0$) when there is a direction $\boldsymbol{c}$ of very small variance in the data ($\sigma \to 0$) and the boundary $\mathcal{C}$ lies close to the data along this direction ($d(\boldsymbol{i},\mathcal{C}) \to 0$).

\item \underline{Minimum strength.} We call the hyperplane of the nearest centroid classifier the \emph{bisecting boundary}, and denote it $\mathcal{B}$. By definition, $\mathcal{B}$ is the unique classification boundary verifying ${\boldsymbol{m}(\boldsymbol{i},\mathcal{B}) = \boldsymbol{j}}$ (we assume that ${\boldsymbol{i} \neq \boldsymbol{j}}$ such that $\mathcal{B}$ is well-defined). Remark that we have, for a classification boundary $\mathcal{C}$:
$${m(I,\mathcal{C}) = J} \implies {\boldsymbol{m}(\boldsymbol{i},\mathcal{C}) = \boldsymbol{j}} \quad \text{but} \quad {\boldsymbol{m}(\boldsymbol{i},\mathcal{B}) = \boldsymbol{j}} \centernot\implies {m(I,\mathcal{B}) = J}$$
Hence, if there exists a classification boundary $\mathcal{C}$ that does not suffer from adversarial examples on $I$, then it is unique and equal to $\mathcal{B}$; but $\mathcal{B}$ can suffer from adversarial examples. In the following, we consider that $\mathcal{B}$ \emph{minimises} the phenomenon of adversarial examples, even when $\mathcal{B}$ does suffer from adversarial examples (see figure~\ref{weakAE}, left). Then, we can say that \emph{the strength of the adversarial examples affecting $I$ is minimised when the distance $\norm{\boldsymbol{j} - \boldsymbol{m}(\boldsymbol{i},\mathcal{C})}$ tends to 0} (see figure~\ref{weakAE}, right).

\begin{figure}[ht]
  \centering
  \includegraphics[width= \textwidth]{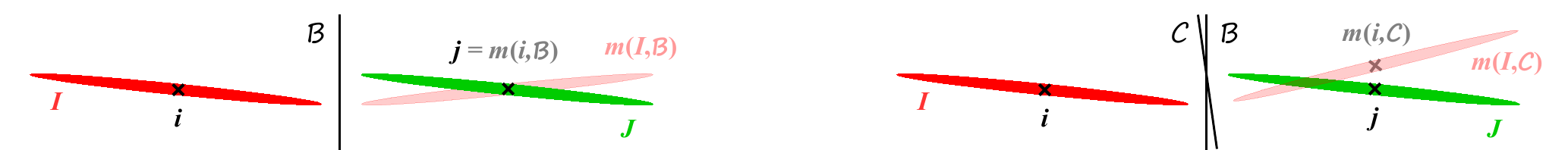}
  \caption{Left: the adversarial examples phenomenon is minimised when $\boldsymbol{j} = \boldsymbol{m}(\boldsymbol{i},\mathcal{B})$ even when $J \neq m(I,\mathcal{B})$. Right: the smaller the distance $\norm{\boldsymbol{j} - \boldsymbol{m}(\boldsymbol{i},\mathcal{C})}$, the weaker the adversarial examples affecting $I$.}
  \label{weakAE}
\end{figure}
\end{description}

Based on the previous considerations, and using the arctangent in order to bound the values in the finite interval $[0,\,\pi/2[$, we formally define the \emph{strength} $s(I,\mathcal{C})$ of the adversarial examples affecting $I$ through $\mathcal{C}$ as:
$$\boxed{s(I,\mathcal{C}) = \arctan\left(\frac{\norm{\boldsymbol{j} - \boldsymbol{m}(\boldsymbol{i},\mathcal{C})}}{\norm{\boldsymbol{i} - \boldsymbol{m}(\boldsymbol{i},\mathcal{C})}}\right)}$$
$s(I,\mathcal{C})$ is maximised at $\pi/2$ when $\norm{\boldsymbol{i} - \boldsymbol{m}(\boldsymbol{i},\mathcal{C})} \to 0$ and minimised at 0 when $\norm{\boldsymbol{j} - \boldsymbol{m}(\boldsymbol{i},\mathcal{C})} \to 0$

\newpage
\subsubsection{The adversarial strength is the deviation angle}

In our analysis, the bisecting boundary $\mathcal{B}$ of the nearest centroid classifier plays a special role: it minimises the strength of the adversarial examples affecting $I$ and $J$. We note $\boldsymbol{b}$ its normal weight vector (we assume that $\norm{\boldsymbol{b}}_2 = 1$) and $b_0$ its bias. Given a classifier $\mathcal{C}$ specified by a normal weight vector $\boldsymbol{c}$ and a bias $c_0$, we call \emph{deviation angle} of $\mathcal{C}$ with regards to $\mathcal{B}$ the angle $\delta_c$ between $\boldsymbol{c}$ and $\boldsymbol{b}$. More precisely, we can express $\boldsymbol{c}$ as a function of $\boldsymbol{b}$, a unit vector orthogonal to $\boldsymbol{b}$ that we note $\boldsymbol{b}^\perp_c$, and the deviation angle $\delta_c$ as:
$$\boldsymbol{c} = \cos(\delta_c)\,\boldsymbol{b} + \sin(\delta_c)\,\boldsymbol{b}^\perp_c$$
We can then derive (see appendix A) the strengths of the adversarial examples affecting $I$ and $J$ through $\mathcal{C}$ in terms of the deviation angle $\delta_c$ and the ratio $r_c = c_0/\norm{\boldsymbol{i}}$ (with the origin $\boldsymbol{0}$ at the midpoint between $\boldsymbol{i}$ and $\boldsymbol{j}$):
$$\boxed{s(I,\mathcal{C}) = \arctan\left(\frac{\sqrt{\sin^2(\delta_c)+r_c^2}}{\cos(\delta_c)+r_c}\right) \quad \text{and} \quad s(J,\mathcal{C}) = \arctan\left(\frac{\sqrt{\sin^2(\delta_c)+r_c^2}}{\cos(\delta_c)-r_c}\right)}$$

\begin{figure}[ht]
  \centering
  \includegraphics[width= \textwidth]{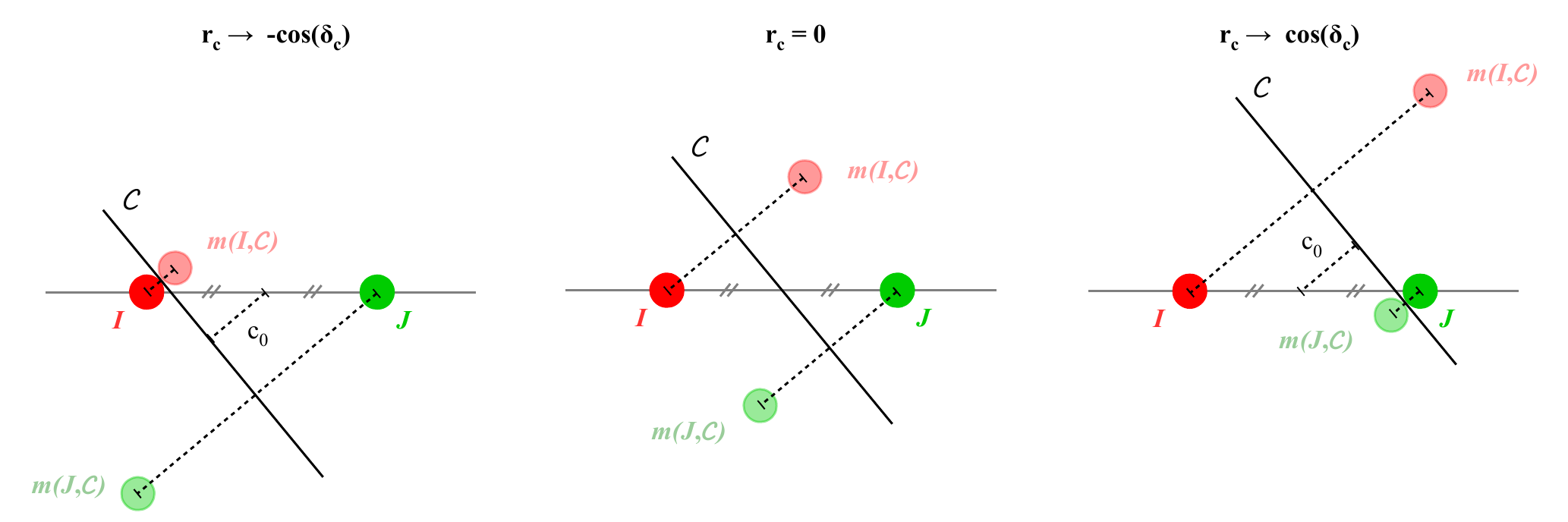}
  \caption{The parameter $r_c$ controls the relative strengths of the adversarial examples affecting $I$ and $J$.}
  \label{rc}
\end{figure}

\begin{description}
\item \underline{Effect of $r_c$:}\\
If we assume that $\mathcal{C}$ separates $\boldsymbol{i}$ and $\boldsymbol{j}$, then we must have $-\cos(\delta_c) < r_c < \cos(\delta_c)$.\\
$\displaystyle \text{When } r_c\to-\cos(\delta_c) \text{, we have: } s(I,\mathcal{C}) \to \pi/2 \quad \text{and} \quad s(J,\mathcal{C}) \to \pi/2 - \arctan(2\cos(\delta_c))$.\\
$\displaystyle \text{When } r_c\to\cos(\delta_c) \text{, we have: } s(I,\mathcal{C}) \to \pi/2 - \arctan(2\cos(\delta_c)), \quad \text{and} \quad s(J,\mathcal{C}) \to \pi/2$.\\[0.1cm]
The parameter $r_c$ controls the relative strengths of the adversarial examples affecting $I$ and $J$. It can lead to strong adversarial examples on one class at a time (see figure~\ref{rc}).
\end{description}

\noindent In the following, we assume that $r_c \approx 0$, so that: 
$$\boxed{s(I,\mathcal{C}) \approx s(J,\mathcal{C}) \approx s(\mathcal{C}) = \arctan\left(\frac{\sqrt{\sin^2(\delta_c)}}{\cos(\delta_c)}\right) = |\delta_c|}$$
In words, when $\mathcal{C}$ passes close to the mean of the classes centroids ($r_c \approx 0$), the strength of the adversarial examples affecting $I$ is approximately equal to the strength of the adversarial examples affecting $J$ and can be reduced to the deviation angle $|\delta_c|$. In that case we can speak of the \emph{adversarial strength} without mentioning the class affected: it is minimised for $\delta_c = 0$ (i.e. $\mathcal{C} \approx \mathcal{B}$) and maximised when $|\delta_c|$ tends to $\pi/2$.

\subsubsection{Boundary tilting and its influence on classification}

In previous sections, we defined the notion of adversarial strength and showed that it can be reduced to the deviation angle between the weight vector $\boldsymbol{c}$ of the classifier considered and the weight vector $\boldsymbol{b}$ of the nearest centroid classifier. Here, we evaluate the effect on the classification performance of tilting the weight vector $\boldsymbol{c}$ by an angle $\theta$ along an arbitrary direction.

\paragraph{}
Let $\boldsymbol{z}$ be a unit vector that we call the \emph{zenith direction}. We can express $\boldsymbol{c}$ as a function of $\boldsymbol{z}$, a unit vector orthogonal to $\boldsymbol{z}$ that we note $\boldsymbol{z}^\perp_c$ and an angle $\theta_c$ that we call the \emph{inclination angle} of $\mathcal{C}$ along $\boldsymbol{z}$:
$$\boldsymbol{c} = \cos(\theta_c)\,\boldsymbol{z}^\perp_c + \sin(\theta_c)\,\boldsymbol{z}$$
We say that we \emph{tilt the boundary} $\mathcal{C}$ along the zenith direction $\boldsymbol{z}$ by an angle $\theta$ when we define a new boundary $\mathcal{C}_\theta$ specified by its normal weight vector $\boldsymbol{c}_\theta$ and its bias $c_{\theta0}$ as follow:
$$\boldsymbol{c}_\theta = \cos(\theta_c + \theta)\,\boldsymbol{z}^\perp_c + \sin(\theta_c + \theta)\,\boldsymbol{z}$$
$$c_{\theta0} = c_0\,\cos(\theta_c + \theta)/\cos(\theta_c)$$
Let $S$ be the set of all the images in $I$ and $J$. Abusing the notation, we refer to the sets of all classification scores through $\mathcal{C}$ and $\mathcal{C}_\theta$ by $d(S,\mathcal{C})$ and $d(S,\mathcal{C}_\theta)$. We can show (see appendix B) that:
$$\boxed{d(S,\mathcal{C}) =  \boldsymbol{u} \cdot P \quad \text{and} \quad d(S,\mathcal{C}_\theta) = \boldsymbol{u}_\theta \cdot P}$$
Where $\boldsymbol{u} = (\cos(\theta_c),\; \sin(\theta_c))$ and $\boldsymbol{u}_\theta = (\cos(\theta_c + \theta),\; \sin(\theta_c + \theta))$ are the unit vectors rotated by the angles $\theta_c$ and $\theta_c + \theta$ relatively to the x-axis and $P = S \cdot (\boldsymbol{z}^\perp_c + c_0/\cos(\theta_c),\; \boldsymbol{z})^\top$ is the projection of $S$ on the plane $(\boldsymbol{z}^\perp_c,\; \boldsymbol{z})$ horizontally translated by $c_0/\cos(\theta_c)$.

\begin{figure}[ht]
  \centering
  \includegraphics[width= 0.5\textwidth]{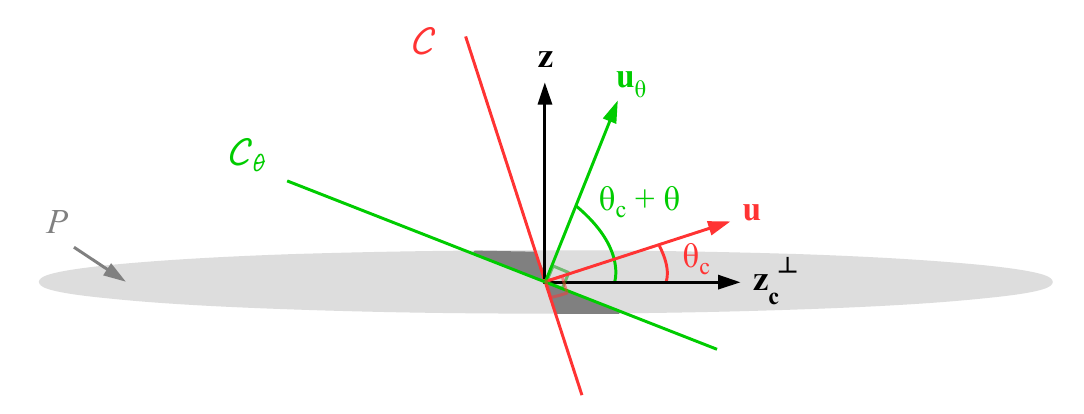}
  \caption{The rate of change $roc(\theta)$ is the proportion of elements in $P$ classified differently by $\mathcal{C}$ and $\mathcal{C}_\theta$ (dark grey area in the figure). It is small as long as the variance of the data in $S$ along $\boldsymbol{z}$ is small and the angle $\theta_c + \theta$ is not too close to $\pi/2$.}
  \label{roc}
\end{figure}

\paragraph{}
Now we define the \emph{rate of change} between $\mathcal{C}$ and $\mathcal{C}_\theta$ and note $roc(\theta)$ the proportion of elements in $S$ that are classified differently by $\mathcal{C}$ and $\mathcal{C}_\theta$ (i.e. the elements $\boldsymbol{x}$ in $S$ for which $\sign(d(\boldsymbol{x},\mathcal{C})) \neq \sign(d(\boldsymbol{x},\mathcal{C}_\theta))$). In general, we cannot deduce a closed-form expression of $roc(\theta)$. However, we can represent it graphically in the plane $(\boldsymbol{z}^\perp_c,\; \boldsymbol{z})$ and we see that $roc(\theta)$ is small as long as the variance of the data in $S$ along the zenith direction $\boldsymbol{z}$ is small and the angle $\theta_c + \theta$ is not too close to $\pi/2$ (see figure~\ref{roc}). 

\noindent Let us note $v_{\boldsymbol{z}}^\perp$ and $v_{\boldsymbol{z}}$ the variances of the data in $S$ along the directions $\boldsymbol{z}^\perp_c$ and $\boldsymbol{z}$ respectively. We present below two situations of interest where $roc(\theta)$ can be expressed in closed-form.
\begin{enumerate}[parsep=0.1cm, itemsep=0.1cm, topsep=0.1cm]
\item When $P$ is flat along the zenith component (i.e. when $v_{\boldsymbol{z}}$ is null), we have:
$$d(S,\mathcal{C}) = \cos(\theta_c)\,(S \cdot \boldsymbol{z}^\perp_c + c_0/\cos(\theta_c)) \quad \text{and} \quad d(S,\mathcal{C}_\theta) = \cos(\theta_c + \theta)\,(S \cdot \boldsymbol{z}^\perp_c + c_0/\cos(\theta_c))$$
Hence:
$$\boxed{d(S,\mathcal{C}_\theta) = \frac{\cos(\theta_c + \theta)}{\cos(\theta_c)}\,d(S,\mathcal{C})}$$
For all $\theta_c + \theta$ in $]-\pi/2,\,\pi/2[$, the sign of $d(S,\mathcal{C}_\theta)$ is equal to the sign of $d(S,\mathcal{C})$: every element of $S$ is classified in the same way by $\mathcal{C}$ and $\mathcal{C}_\theta$ and $roc(\theta) = 0$.\\[0.1cm]
\textit{When the variance along the zenith direction is null, the classification of the elements in $S$ is unaffected by the tilting of the boundary.}
\item When $P$ follows a bivariate normal distribution $\mathcal{N}(\boldsymbol{0},\boldsymbol{\Sigma})$ with $\boldsymbol{\Sigma} = \diag(v_{\boldsymbol{z}}^\perp, \, v_{\boldsymbol{z}})$, then we can show (see appendix C) that:
$$\boxed{roc(\theta) = \frac{1}{\pi}\left[\arctan\left(\sqrt{\frac{v_{\boldsymbol{z}}}{v_{\boldsymbol{z}}^\perp}}\tan(x)\right)\right]_{\theta_c}^{\theta_c+\theta}}$$

For instance if $v_{\boldsymbol{z}}^\perp = 1$ and $v_{\boldsymbol{z}} = 10^{-6}$, and the boundaries $\mathcal{C}$ and $\mathcal{C}_\theta$ are tilted at $10\%$ and $90\%$ respectively along $\boldsymbol{z}$ ($\theta_c = 0.1\,\pi/2$ and $\theta_c+\theta = 0.9\,\pi/2$)), then we have $roc(\theta) = 0.2\%$.\\[0.1cm]
\textit{When the variance along the zenith direction is small enough, the classification of the elements in $S$ is very lightly affected by the tilting of the boundary.}
\end{enumerate}

\subsubsection{Boundary tilting at the origin of strong adversarial examples}

Finally, we show that the boundary tilting mechanism can lead to the existence of strong adversarial examples, without affecting the classification performance. 

\paragraph{}
Imagine that we choose the zenith direction $\boldsymbol{z}$ orthogonal to $\boldsymbol{b}$. Then we can express $\boldsymbol{z}^\perp_c$ as a function of $\boldsymbol{b}$, a unit vector orthogonal to $\boldsymbol{b}$ (and $\boldsymbol{z}$) that we note $\boldsymbol{y}_c$ and an angle $\phi_c$ that we call the \emph{azimuth angle} of $\mathcal{C}$ with regards to $\boldsymbol{z}$ and $\boldsymbol{b}$:
$$\boldsymbol{c} = \cos(\theta_c)\left[\,\cos(\phi_c)\,\boldsymbol{b} + \sin(\phi_c)\,\boldsymbol{y}_c\,\right] + \sin(\theta_c)\,\boldsymbol{z}$$
Now, imagine that we tilt the boundary $\mathcal{C}$ along the zenith direction $\boldsymbol{z}$ while keeping the azimuth angle $\phi_c$ constant. We can express the weight vector $\boldsymbol{c}_\theta$ of the tilted boundary $\mathcal{C}_\theta$ both as a function of its inclination angle $\theta_c + \theta$ and the azimuth angle $\phi_c$, and as a function of its deviation angle $\delta_c+\delta$:
$$\boldsymbol{c}_\theta = \cos(\theta_c+\theta)\left[\,\cos(\phi_c)\,\boldsymbol{b} + \sin(\phi_c)\,\boldsymbol{y}_c\,\right] + \sin(\theta_c+\theta)\,\boldsymbol{z} \quad \text{and} \quad \boldsymbol{c}_\theta = \cos(\delta_c+\delta)\,\boldsymbol{b} + \sin(\delta_c+\delta)\,\boldsymbol{b}^\perp_c$$

\noindent We see that the deviation angle $\delta_c+\delta$ of $\mathcal{C}_\theta$ depends on the inclination angle $\theta_c+\theta$ and the azimuth angle $\phi_c$:
$$\boxed{\cos(\delta_c+\delta) = \cos(\theta_c+\theta)\,\cos(\phi_c)}$$

\begin{figure}[ht]
  \centering
  \includegraphics[width= 0.5\textwidth]{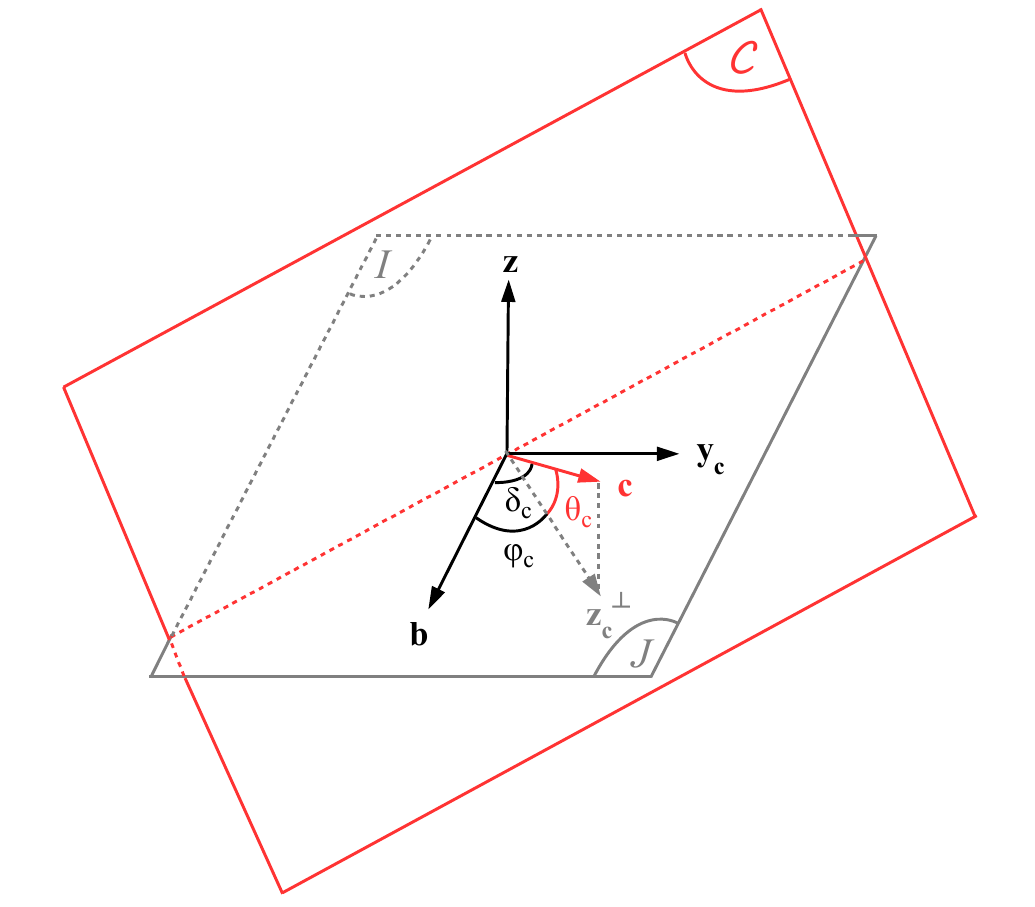}
  \caption{Illustration in 3 dimensions of the relationship between the deviation angle $\delta_c$, the inclination angle $\theta_c$ and the azimuth angle $\phi_c$. When the variance $v_{\boldsymbol{z}}$ is null and the azimuth angle $\phi_c$ is kept constant, it is possible to have the deviation angle $\delta_c$ approaching $\pi/2$ (resulting in strong adversarial examples) by tilting along the direction $\boldsymbol{z}$ without affecting the classification performance.}
  \label{tilting}
\end{figure}

\paragraph{}
In order for $\mathcal{C}_\theta$ to suffer  from strong adversarial examples (i.e. $|\delta_c+\delta| \to \pi/2$), it is sufficient to tilt along a zenith direction $\boldsymbol{z}$ orthogonal to $\boldsymbol{b}$ (i.e. $|\theta_c+\theta| \to \pi/2$). If in addition the direction $\boldsymbol{z}$ is such that the variance $v_{\boldsymbol{z}}$ is small, then the rate of change $roc(\theta)$ will be small and the classification boundaries $\mathcal{C}$ and $\mathcal{C}_\theta$ will perform similarly (when $v_{\boldsymbol{z}} = 0$, $\mathcal{C}$ and $\mathcal{C}_\theta$ perform exactly in the same way: see figure~\ref{tilting}).

\begin{quote}
\textit{For any classification boundary $\mathcal{C}$, there always exist a tilted boundary $\mathcal{C}_\theta$ such that $\mathcal{C}$ and $\mathcal{C}_\theta$ perform in the same way $(v_{\boldsymbol{z}} = 0)$ or almost in the same way $(0 < v_{\boldsymbol{z}} \ll 1)$, and $\mathcal{C}_\theta$ suffers from adversarial examples of arbitrary strength (as long as there are directions of low variance in the data).}
\end{quote}

\subsubsection{Taxonomy of adversarial examples}

\begin{figure}[ht]
  \centering
  \includegraphics[width= 0.66\textwidth]{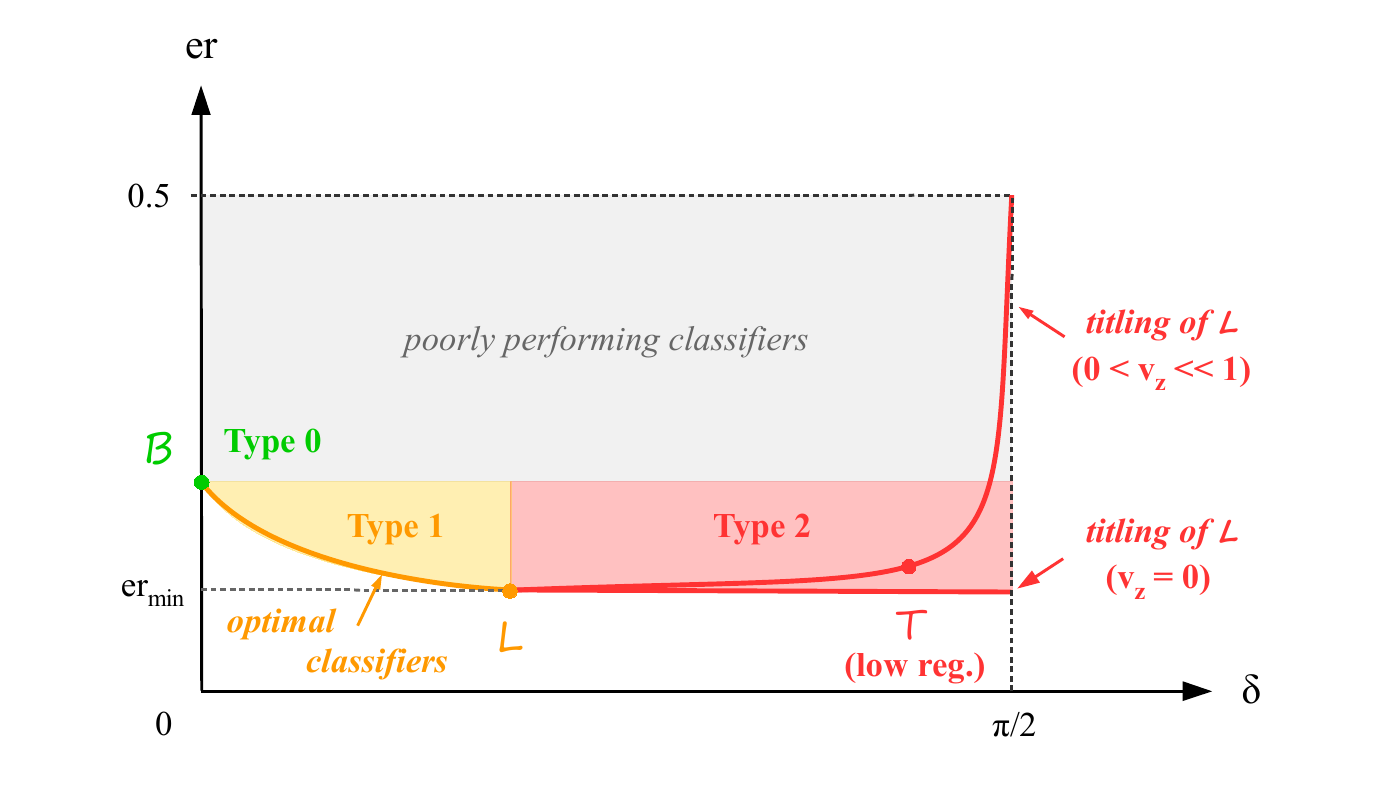}
  \caption{Deviation Angle - Error Rate diagram. The position of the optimal classifiers, including in particular the bisecting boundary $\mathcal{B}$ and the label boundary $\mathcal{L}$, is indicated. The effect of tilting $\mathcal{L}$ along a direction of no variance ($v_{\boldsymbol{z}} = 0$) or low variance ($0 < v_{\boldsymbol{z}} \ll 1$), is also illustrated. This mechanism results in a training boundary $\mathcal{T}$ that suffers from strong type 2 adversarial examples when the level of regularisation used is low.}
  \label{types}
\end{figure}

Given a classifier $\mathcal{C}$, we note $\delta(\mathcal{C})$ its deviation angle and $er(\mathcal{C})$ its error rate on $S$. In the following, we analyse the distribution of all linear classifiers in the \emph{deviation angle - error rate diagram}. To start with, we consider the nearest centroid classifier $\mathcal{B}$ as a baseline and discard all classifiers with an error rate superior to $er(\mathcal{B})$ as poorly performing. We also note $er_\text{min}$ the minimum error rate achievable on $S$ (in general, $er_\text{min} < er(\mathcal{B})$). For a given error rate comprised between $er(\mathcal{B})$ and $er_\text{min}$, we say that a classifier is \emph{optimal} if it minimises the deviation angle. In particular, we call \emph{label boundary} and we note $\mathcal{L}$ the optimal classifier verifying $er(\mathcal{L}) = er_\text{min}$. In the deviation angle - error rate diagram, the set of optimal classifiers forms a strictly decreasing curve segment connecting $\mathcal{B}$ (minimising the strength of the adversarial examples) to $\mathcal{L}$ (minimising the error rate). Any classifier with a deviation angle greater than $\delta(\mathcal{L})$ is then necessarily suboptimal: there is always another classifier performing at least as well and suffering from weaker adversarial examples (see figure~\ref{types}). 

\bigskip
\noindent Based on these considerations, we propose to define the following taxonomy:
\begin{description}[parsep=0.1cm, itemsep=0cm, topsep=0.1cm]
\item[Type 0:] adversarial examples affecting $\mathcal{B}$. They \emph{minimise} the phenomenon of adversarial examples.
\item[Type 1:] adversarial examples affecting the classifiers $\mathcal{C}$ such that $0 \leq \delta(\mathcal{C}) \leq \delta(\mathcal{L})$. They affect in particular the \emph{optimal classifiers}. The inconvenience of their existence is balanced by the performance gains allowed.
\item[Type 2:] adversarial examples affecting the classifiers $\mathcal{C}$ such that $\delta(\mathcal{L}) < \delta(\mathcal{C})$. They only affect \emph{suboptimal classifiers} resulting from the tilting of optimal classifiers along directions of low variance.
\end{description}

\paragraph{}
Let us call \emph{training boundary} and note $\mathcal{T}$ the boundary defined by a standard classification method such as SVM or logistic regression. In practice, $I$ and $J$ are unlikely to be mirror classes of each other through $\mathcal{B}$ and hence $\mathcal{T}$ is expected to at least suffer from type 0 adversarial examples. In fact, $\mathcal{B}$ is also unlikely to minimise the error rate on $S$ and if $\mathcal{T}$ performs better than $\mathcal{B}$, then $\mathcal{T}$ is also expected to suffer from type 1 adversarial examples. Note that there is no restriction in theory on $\delta(\mathcal{L})$ and on some problems, type 1 adversarial examples can be very strong. However, $\mathcal{T}$ is a priori not expected to suffer from type 2 adversarial examples: why would SVM or logistic regression define a classifier that is suboptimal in such a way? In the following two sections, we show experimentally with SVM that the regularisation level plays a crucial role in controlling the deviation angle of $\mathcal{T}$. When the regularisation level is very strong (i.e. when the SVM margin contains all the data), $\mathcal{T}$ converges towards $\mathcal{B}$ and the deviation angle is null. When SVM is correctly regularised, $\mathcal{T}$ is allowed to deviate from $\mathcal{B}$ sufficiently to converge towards $\mathcal{L}$: the optimal classifier minimising the error rate. However when the regularisation level is too low, the inclination of $\mathcal{T}$ along directions of low variance ends up overfitting the training data, resulting in the existence of strong type 2 adversarial examples.

\newpage
\subsection{Return to the toy problem}
\label{sec: Return to the toy problem}

\begin{figure}[p]
  \centering
  \includegraphics[width=\textwidth]{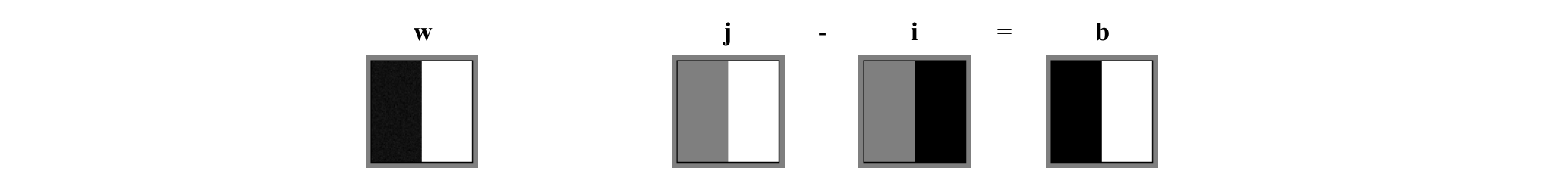}
  \caption{The weight vector $\boldsymbol{w}$ obtained using SVM in figure~\ref{toyExample} is equal to the weight vector $\boldsymbol{b}$ of the nearest centroid classifier, obtained by subtracting the mean image $\boldsymbol{i}$ of the class $I$ to the mean image $\boldsymbol{j}$ of the class $J$.}
  \label{centroids}
\end{figure}

In light of the mathematical analysis presented in the previous sections, we now return to the toy problem introduced in section~\ref{sec: Linear behaviour is not sufficient to cause adversarial examples} (see figure~\ref{toyExample}). Firstly, we can confirm that the boundary defined by SVM satisfies the condition we gave for the non-existence of adversarial examples: the weight vector $\boldsymbol{w}$ is equal to the weight vector $\boldsymbol{b}$ of the nearest centroid classifier $\mathcal{B}$ (see figure~\ref{centroids}) and we have $m(I,\mathcal{B}) = J$ and $m(J,\mathcal{B}) = I$. Indeed, mirroring  an image that belongs to $I$ through $\mathcal{B}$ changes the colour of its right half image from black to white and results in an image that belongs to $J$ (and conversely).

\begin{figure}[p]
  \centering
  \includegraphics[width=\textwidth]{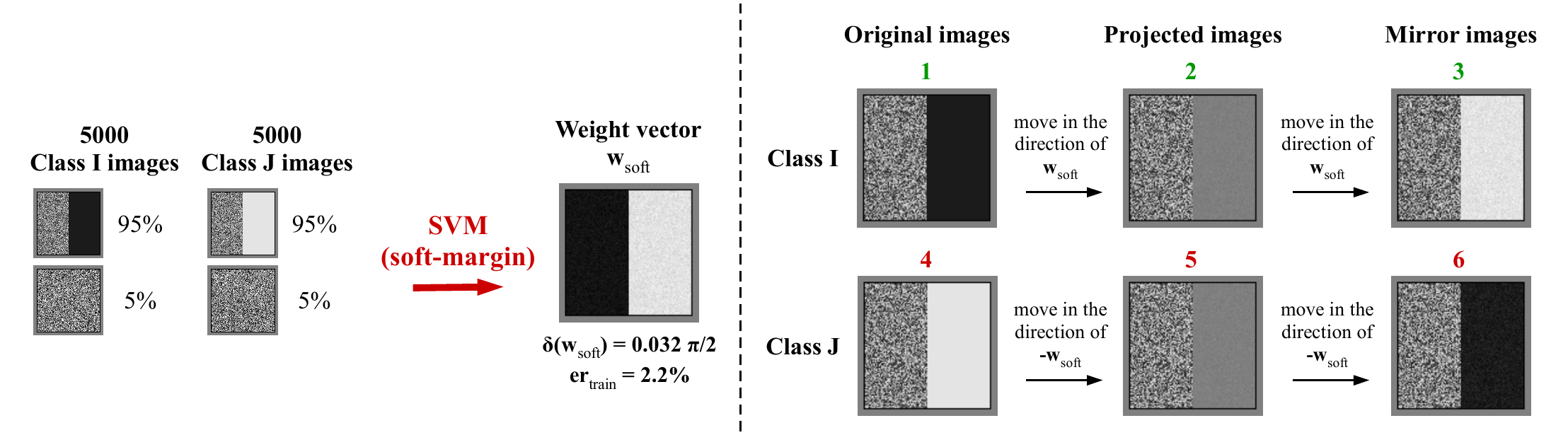}
  \caption{Left: toy problem where 5\% of the data is corrupted to purely random images such that the two classes are not linearly separable ($er_\text{min} = 2.5\%$). With a proper level of regularisation (soft-margin), the training data is only slightly overfitted ($er_\text{train} = 2.2\%$) and the weight vector $\boldsymbol{w}_\text{soft}$ defined by SVM only deviates slightly from $\boldsymbol{b}$ ($\delta(\boldsymbol{w}_\text{soft}) = 0.032\,\pi/2$). Right: as a result, adversarial examples are very weak.}
  \label{withRegularisation}
\end{figure}

\begin{figure}[p]
  \centering
  \includegraphics[width=\textwidth]{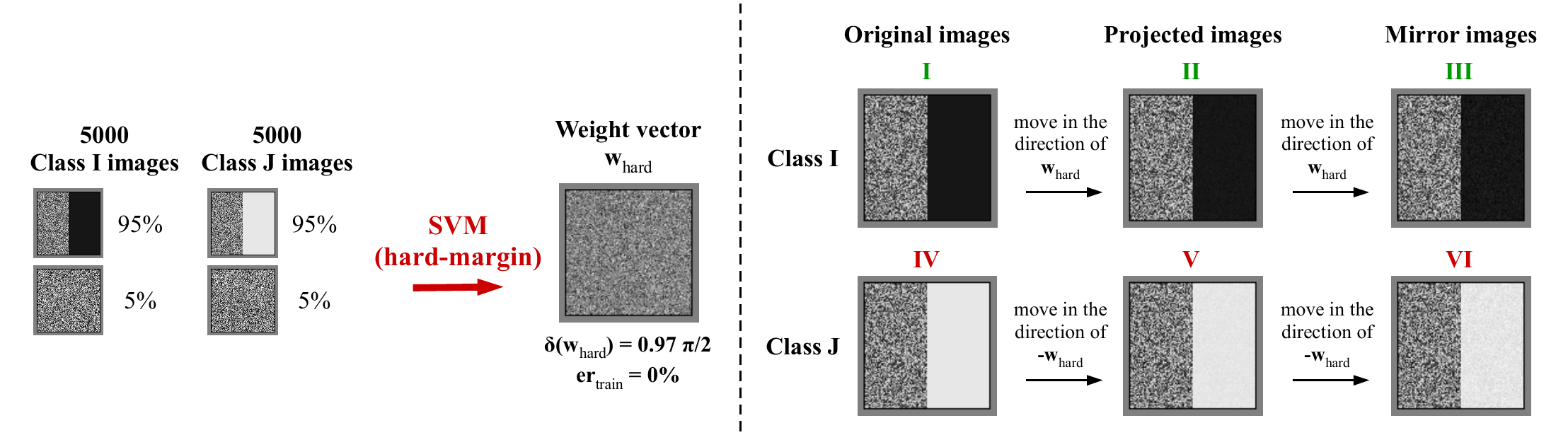}
  \caption{Left: same toy problem as before. Without regularisation (with hard-margin), the training data is entirely overfitted ($er_\text{train} = 0\%$) and the weight vector $\boldsymbol{w}_\text{hard}$ defined by SVM deviates from $\boldsymbol{b}$ considerably ($\delta(\boldsymbol{w}_\text{hard}) = 0.97\,\pi/2$). Right: as a result, adversarial examples are very strong.}
  \label{withoutRegularisation}
\end{figure}

\paragraph{}
Secondly, we can illustrate the effect of the regularisation level used on the deviation angle (and hence on the adversarial strength). To start with, we modify the toy problem such that $er_\text{min} > 0$ (when $er_\text{min} = 0$, overfitting is not likely to happen). We do this by corrupting 5\% of the images in $I$ and $J$ into fully randomised images, such that $er_\text{min} = 2.5\%$ (half of the corrupted data is necessarily misclassified). Note that on this problem, $er(\mathcal{B}) = er_\text{min}$, hence $\mathcal{B} = \mathcal{L}$ and $\mathcal{B}$ is the only optimal classifier. When we perform SVM with regularisation (soft-margin), we obtain a weight vector $\boldsymbol{w}_\text{soft}$ approximately equal to $\boldsymbol{b}$ (see figure~\ref{withRegularisation}). The small deviation can be explained by the fact that the training data has been slightly overfitted (the training error is $2.2\% < er_\text{min}$) and corresponds to very weak adversarial examples. Without regularisation however (hard-margin), the deviation of the weight vector $\boldsymbol{w}_\text{hard}$ is very strong (see figure~\ref{withoutRegularisation}). In that case, the training data is completely overfitted (the training error is 0\%), resulting in the existence of strong type 2 adversarial examples. Interestingly, these adversarial examples possess the same characteristics as the ones observed with GoogLeNet on ImageNet in \citep{goodfellow2014explaining} --- the perturbation is barely perceptible, high-frequency and cannot be meaningfully interpreted --- even though the classifier is linear.

\begin{figure}[p]
  \centering
  \includegraphics[width=\textwidth]{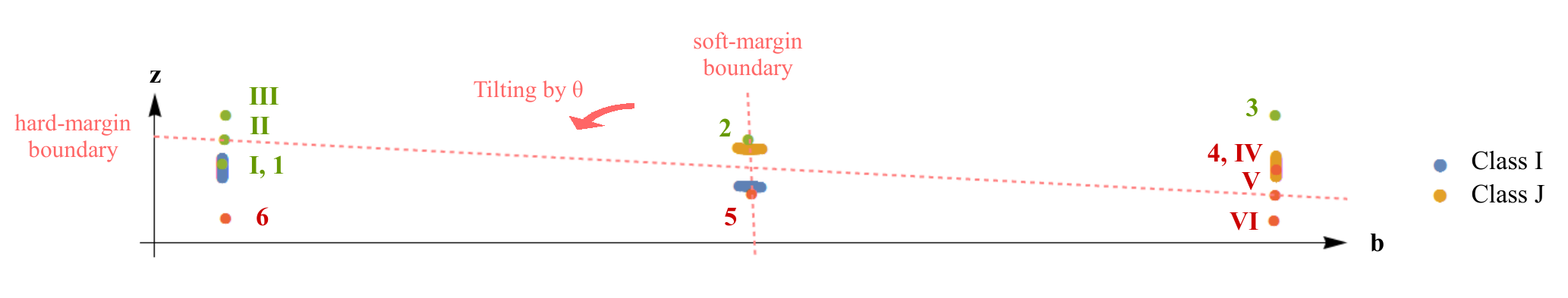}
  \caption{Projection of the training data in the plane $(\boldsymbol{b},\boldsymbol{z})$ where $\boldsymbol{z} = \text{normalise}(\boldsymbol{w}_\text{hard} - (\boldsymbol{w}_\text{hard} \cdot \boldsymbol{b})\,\boldsymbol{b})$. The images in $I$ appear on the left, the images in $J$ appear on the right, and the corrupted images appear in the middle. The soft-margin and hard-margin boundaries are drawn as dashed red lines. Note that the hard-margin boundary overfits the training data by finding a direction that separates the corrupted data completely (this separation does not generalise to novel test data). The positions of the original images, projected images and mirror images of the figures~\ref{withRegularisation}~and~\ref{withoutRegularisation} are also shown: the adversarial examples III and VI of the hard-margin boundary are much closer to their respective original images than the adversarial examples 3 and 6 of the soft-margin boundary.}
  \label{projections}
\end{figure}

\paragraph{}
Finally, we can visualise the boundary tilting mechanism by plotting the projections of the data on the plane $(\boldsymbol{b},\boldsymbol{z})$, where $\boldsymbol{z}$ is the zenith direction along which $\boldsymbol{w}_\text{hard}$ is tilted (see figure~\ref{projections}). We observe in particular how the overfitting of the corrupted data leads to the existence of the strong type 2 adversarial examples: maximising the \emph{minimal} separation of the two classes (the margin) results in a very small \emph{average} separation (making adversarial examples possible). This effect is very reminiscent of the \emph{data piling} phenomenon studied by \cite{marron2007distance} and \cite{ahn2010maximal} on high-dimension low-sample size data.

\subsection{Return to MNIST}
\label{sec: Return to MNIST}

\begin{figure}[p]
  \centering
  \includegraphics[width=\textwidth]{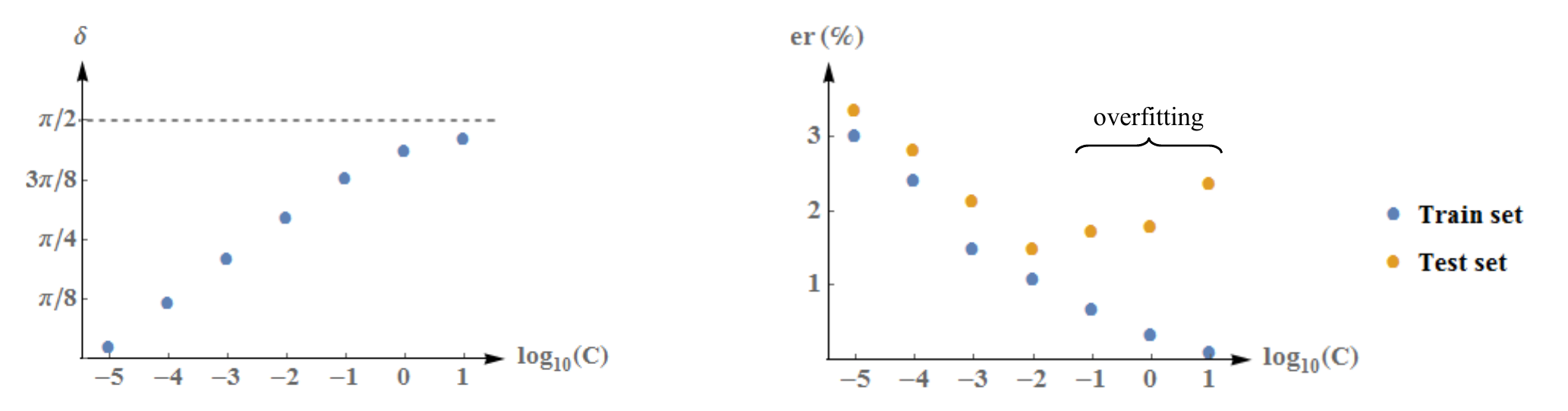}
  \caption{Left: the deviation angle of the weight vector defined by SVM increases almost linearly with the $\log_{10}$ of the soft-margin parameter $C$. Right: The error rate on training data decreases with $\log_{10}(C)$. The error rate on test data is minimised for an intermediate level of regularisation ($\log_{10}(C) = -2$) and overfitting happens for low levels of regularisation ($\log_{10}(C) = -1, 0$ and $1$).}
  \label{deviationAngleErrorRate}
\end{figure}

We now revisit the 3s vs 7s MNIST problem. In particular, we study the effect of varying the regularisation level by performing SVM classification with seven different values for the soft-margin parameter: ${\log_{10}(C) = -5, -4, -3, -2, -1, 0}$ and $1$. The first remark we can make is that there is a strong, direct correlation between the deviation angle of the weight vector defined by SVM and the regularisation level used (see figure~\ref{deviationAngleErrorRate}, left). When regularisation is high (i.e. when $C$ is low), the SVM weight vector is very close to the weight vector of the nearest centroid classifier $\boldsymbol{b}$ ($\delta = 0.048\,\pi/2$). Conversely when regularisation is low (i.e. when $C$ is high), the SVM weight vector is almost orthogonal to $\boldsymbol{b}$ ($\delta = 0.92\,\pi/2$). As expected, the error rate on test data is minimised for an intermediate level of regularisation and overfitting happens for low regularisation: for $\log_{10}(C) = -1, 0$ and $1$, the error rate on training data approaches 0\% while the error rate on test data increases (see figure~\ref{deviationAngleErrorRate}, right).

\begin{figure}[p]
  \includegraphics[width=\textwidth]{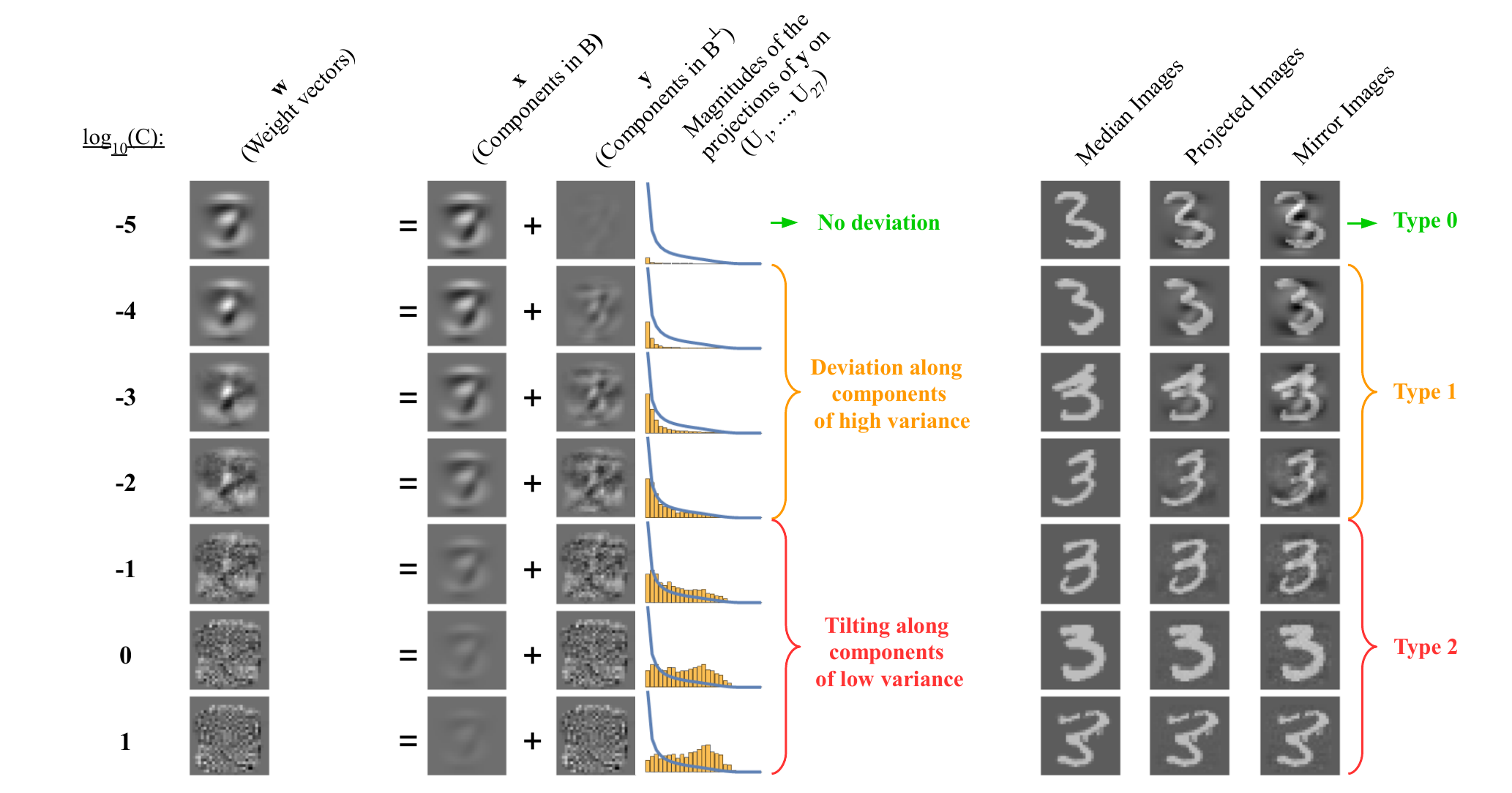}
  \caption{Left: weight vector $\boldsymbol{w}$ defined by SVM for different levels of regularisation (controlled with the soft-margin parameter $C$). Middle: decomposition of $\boldsymbol{w}$ into a component $\boldsymbol{x}$ in $B$ and a component $\boldsymbol{y}$ in $B^\perp$. The orange bar charts represent the magnitudes of the projections of $\boldsymbol{y}$ on the subspaces of decreasing variances $U_1, ..., U_{27}$ and the blue curves represent the square root of the total variance in each subspace. Right: Median 3, its projected image and its mirror image for each regularisation level.}
  \label{deviationAndAE}
\end{figure}

\paragraph{}
When we look at the SVM weight vector $\boldsymbol{w}$ for the different levels of regularisation (see figure~\ref{deviationAndAE}, left), we see that it initially resembles the weight vector of the nearest centroid classifier ($\log_{10}(C) = -5$), then deviates away into relatively low frequency directions ($\log_{10}(C) = -4, -3$ and $-2$) before deviating into higher frequency directions, resulting in a ``random noise aspect'', when the training data starts to be overfitted ($\log_{10}(C) = -1, 0$ and $1$). Let us consider $B$ the one-dimensional subspace of $\mathbb{R}^{784}$ generated by $\boldsymbol{b}$, and $B^\perp$ the 783-dimensional subspace of $\mathbb{R}^{784}$, orthogonal complement of $B$. We note $X_\text{train}$ and $Y_\text{train}$ the projections of the training set $S_\text{train}$ on $B$ and $B^\perp$ respectively and we perform a principal component analysis of $Y_\text{train}$, resulting in the 783 principal vectors $\boldsymbol{u}_1$, ..., $\boldsymbol{u}_{783}$. Then, we decompose $B^\perp$ into 27 subspaces $U_1, ..., U_{27}$ of 29 dimensions each, such that $U_1$ is generated by $\boldsymbol{u}_1, ..., \boldsymbol{u}_{29}$, $U_2$ is generated by $\boldsymbol{u}_{30}, ..., \boldsymbol{u}_{58}$, ..., and $U_{27}$ is generated by $\boldsymbol{u}_{755}, ..., \boldsymbol{u}_{783}$. For each weight vector $\boldsymbol{w}$, we decompose it into a component $\boldsymbol{x}$ in $B$ and a component $\boldsymbol{y}$ in $B^\perp$ and we project $\boldsymbol{y}$ on each subspace $U_1$, ..., $U_{27}$ (see figure~\ref{deviationAndAE}, middle). The norms of the projections of $\boldsymbol{y}$ are shown as orange bar charts and the square roots of the total variances in each subspace $U_1$, ..., $U_{27}$ are shown as blue curves. We see that for $\log_{10}(C) = -4, -3$ and $-2$, $\boldsymbol{y}$ is dominated by components of high variance, while for $\log_{10}(C) = -1, 0$ and $1$, $\boldsymbol{y}$ starts to be more dominated by components of low variance: this result confirms that overfitting happens by the tilting of the boundary along components of low variance. Note that $\boldsymbol{w}$ never tilts along flat directions of variation (corresponding to the subspaces $U_{23}, ..., U_{27}$) because for overfitting to take place, there needs to be some variance in the tilting direction. Interestingly, optimal classification seems to happen when each direction is used proportionally to the amount of variance it contains: for $\log_{10}(C) = -2$, the bar chart follows the blue curve faithfully. Finally, we can look at the adversarial examples affecting each weight vector (see figure~\ref{deviationAndAE}, right). In particular, we look at the images of 3s in the test set that are at a median distance from each boundary (median images). We see that the mirror images are closer to their respective original images when the regularisation level is low, resulting in stronger adversarial examples. For $\log_{10}(C) = -5$, the deviation angle is almost null and we can say that the corresponding adversarial example is of type 0. For $\log_{10}(C) = -4, -3$ and $-2$, the increase in deviation angle is associated with an increase in performance and we can say that the corresponding adversarial examples are of type 1. However, for $\log_{10}(C) = -1, 0$ and $1$, the increase in deviation angle only results in overfitting, and we can say that the corresponding adversarial examples are of type 2.

\begin{figure}[p]
  \centering
  \includegraphics[width=\textwidth]{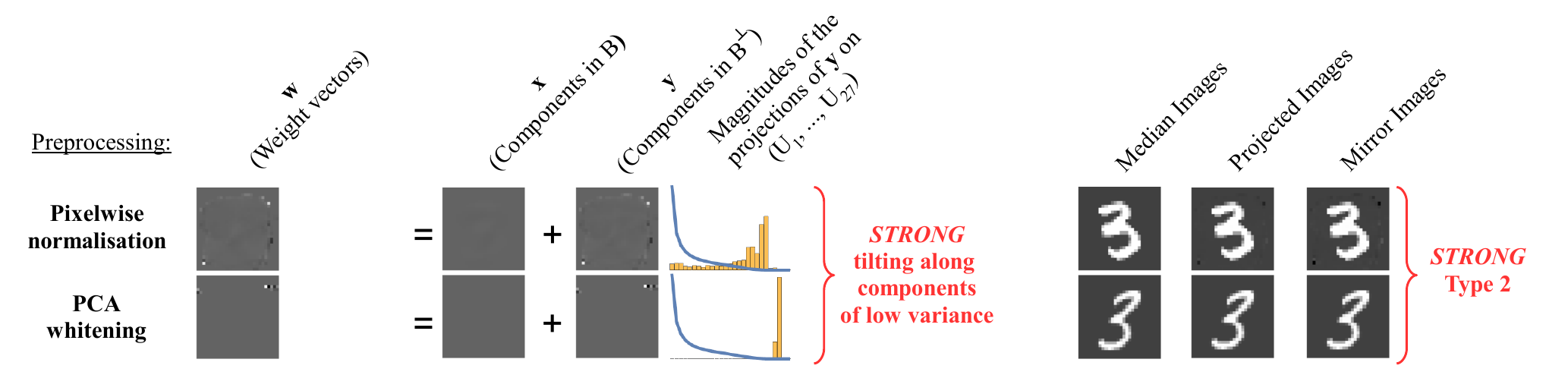}
  \caption{Left: weight vector $\boldsymbol{w}$ defined by SVM with soft-margin after two standard preprocessing methods: pixelwise normalisation and PCA whitening (projected back in pixel space). Middle: decomposition of $\boldsymbol{w}$ into a component $\boldsymbol{x}$ in $B$ and a component $\boldsymbol{y}$ in $B^\perp$. Right: Median 3, its projected image and its mirror image for the two weight vectors.}
  \label{PreprocessingDeviationAndAEV}
\end{figure}

\paragraph{}
These type 2 adversarial examples, like those found on the toy problem, have similar characteristics to the ones affecting GoogLeNet on ImageNet (the adversarial perturbation is barely perceptible and high-frequency). Hence we may hypothesize that the adversarial examples affecting deep networks are also of type 2, originating from a non-linear equivalent of boundary-tilting and caused by overfitting. If this hypothesis is correct, then these adversarial examples might also be fixable by using adapted regularisation. Unfortunately, straightforward l2 regularisation only works when the classification method operates on pixel values: as soon as the regularisation term is applied in a feature space that does not directly reflect pixel distance, it does not effectively prevent the existence of type 2 adversarial examples any more. We illustrate this by performing linear SVM with soft-margin regularisation after two different standard preprocessing methods: pixelwise normalisation and PCA whitening. In the two cases, the soft-margin parameter $C$ is chosen such that the performance is maximised, resulting in a slight boost in performance both for pixelwise normalisation ($er_\text{test} = 1.2\%$) and for PCA whitening ($er_\text{test} = 1.5\%$). Since the preprocessing steps are linear transformations, we can then project the weight vectors obtained back into the original pixel space. We get a deviation angle for the weight vector defined after pixelwise normalisation that is stronger than that of any weight vector defined without preprocessing ($\delta = 0.95\,\pi/2$) and a deviation angle for the weight vector defined after PCA whitening that appears orthogonal to $\boldsymbol{b}$ ($\delta = 1.00\,\pi/2$). The two weight vectors (see figure~\ref{PreprocessingDeviationAndAEV}, left) have a very peculiar aspect: both are strongly dominated by a few pixels, in the periphery of the image for the weight vector defined after pixelwise normalisation and in the top right corner for the weight vector defined after PCA whitening. When we look at the magnitudes of the projections of the $\boldsymbol{y}$ components on the subspaces $U_1, ..., U_{27}$, we see that the dominant pixels correspond to the components where the variance of the data is smallest but non-null (see figure~\ref{PreprocessingDeviationAndAEV}, middle). Effectively, the rescaling of the components of very low variance puts a disproportionate weight on them, forcing the boundary to tilt very significantly. The phenomenon is particularly extreme with PCA whitening where due to numerical approximations, some residual variance was found in components that were not supposed to contain any, and ended up strongly dominating the weight vector\footnote{This effect could be avoided by putting a threshold on the minimum variance necessary before rescaling, as is sometimes done in practice.}. The resulting adversarial examples are unusual (see figure~\ref{PreprocessingDeviationAndAEV}, right). For the pixelwise normalisation preprocessing step, it is possible to change the class of an image by altering the value of pixels that do not affect the digit itself. For the PCA whitening preprocessing step, the perturbation is absolutely non-perceptible: the pixel distance between the original image and the corresponding adversarial example is in the order of $10^{-18}$. With such a small distance, classification is now very sensitive to any perturbation, whether it is adversarial or random (despite this obvious weakness, this classifier performs very well on normal data).

\newpage
\section{Conclusion}

\paragraph{}
This paper contributes to the understanding of the adversarial example phenomenon in several different ways. It introduces in particular:
\begin{description}[parsep=0.1cm, itemsep=0cm, topsep=0.1cm]
\item[A new perspective.] The phenomenon is captured in one intuitive picture: a submanifold of sampled data, intersected by a class boundary lying close to it, suffers from adversarial examples.
\item[A new formalism.] In linear classification, we proposed a strict condition for the non-existence of adversarial examples. We defined adversarial examples as elements of the mirror class and introduced the notion of adversarial strength. Given a classification boundary $\mathcal{C}$, we showed that the adversarial strength can be measured by the deviation angle between $\mathcal{C}$ and the bisecting boundary $\mathcal{B}$ of the nearest centroid classifier. We also defined the boundary tilting mechanism, and showed that there always exists a tilted boundary $\mathcal{C}_\theta$ such that $\mathcal{C}$ and $\mathcal{C}_\theta$ perform in very similar ways, and $\mathcal{C}_\theta$ suffers from adversarial examples of arbitrary strength (as long as there are directions of low variance in the data).
\item[A new taxonomy.] These results led us to define the notion of optimal classifier, minimising the deviation angle for a given error rate. $\mathcal{B}$ is the optimal classifier minimising the adversarial strength and we called label boundary $\mathcal{L}$ the optimal classifier minimising the error rate. When $\mathcal{C} = \mathcal{B}$ and the two classes of images are not mirror classes of each other, we say that $\mathcal{C}$ suffers from adversarial examples of type~0. When the error rate of $\mathcal{C}$ is strictly inferior to the error rate of $\mathcal{B}$, the deviation angle of $\mathcal{C}$ is necessarily strictly positive; as long as it stays inferior to the deviation angle of $\mathcal{L}$, we say that $\mathcal{C}$ suffers from adversarial examples of type~1. When the deviation angle of $\mathcal{C}$ is superior to the deviation angle of $\mathcal{L}$, $\mathcal{C}$ is necessarily suboptimal. In that case we say that $\mathcal{C}$ suffers from adversarial examples of type~2. 
\item[New experimental results.] We introduced a toy problem that does not suffer from adversarial examples, and presented a minimal set of conditions to provoke the apparition of strong type 2 adversarial examples on it. We also showed on the 3s vs 7s MNIST problem that in practice, the regularisation level used plays a key role in controlling the deviation angle, and hence the type of adversarial examples obtained. Type~2 adversarial examples in particular, can be avoided by using a proper level of regularisation. However, we showed that l2 regularisation only helps when it is applied directly in pixel space.
\end{description}

\paragraph{}
An important distinction must be drawn between the different types of adversarial examples. On the one hand, type 0 and type 1 adversarial examples originate from a lack of expressiveness of linear models: their adversarial perturbations do not correspond to the true features disentangling the classes of images, but they can be interpreted (as optimal linear features). On the other hand, type 2 adversarial examples originate from overfitting: their adversarial perturbations are high frequency and largely meaningless (with a characteristic ``random noise aspect''). Due to their similarity with the type 2 adversarial examples affecting linear classifiers, we hypothesised that the adversarial examples affecting state-of-the-art neural networks are also of type 2, symptomatic of overfitting and resulting from a non-linear equivalent of boundary tilting. Unfortunately, we do not know how to effectively regularise deep networks yet. In fact, we do not know whether it is possible to regularise them at all. Neural networks typically operate in a regime where the number of learnable parameters is higher than the number of training images and one could imagine that such models are fundamentally vulnerable to adversarial examples. Perhaps, the adversarial examples phenomenon is to neural systems what Loschmidt's paradox is to statistical physics: a theoretical aberration of extremely low probability in practice. When Loschmidt pointed out that it is possible to create a system that contradicts the second law of thermodynamics (stating that the entropy of a closed system must always increase) by taking an existing closed system and reversing the motion direction of all the particles constituting it, Boltzmann is reported to have answered: ``Go ahead, reverse them!''. Similarly, one could then reply to those who worry about the possible existence of adversarial examples in humans: ``Go ahead, generate them!''.

\newpage
\section*{Appendix}

\subsubsection*{A \quad Expression of the adversarial strength as a function of the deviation angle}

By choosing the origin $\boldsymbol{0}$ at the midpoint between $\boldsymbol{i}$ and $\boldsymbol{j}$, we can ensure that $\boldsymbol{b} = -\boldsymbol{i}/\norm{\boldsymbol{i}} = \boldsymbol{j}/\norm{\boldsymbol{j}}$ and $b_0 = 0$. We then have:
\vspace{-0.2cm}
\begin{flalign*}
\norm{\boldsymbol{i} - \boldsymbol{m}(\boldsymbol{i},\mathcal{C})} & = \norm{\boldsymbol{i} - \boldsymbol{i} + 2\,d(\boldsymbol{i},\mathcal{C})\,\boldsymbol{c}} &&\\
& = 2\,|d(\boldsymbol{i},\mathcal{C})| &&\\
& = 2\,|\boldsymbol{i} \cdot \boldsymbol{c} + c_0| &&\\[-0.5cm]
& = 2\,|\cos(\delta_c)(\boldsymbol{i} \cdot \boldsymbol{b}) + \sin(\delta_c)\,(\overbrace{\boldsymbol{i} \cdot \boldsymbol{b}^\perp_c}^{0}) + c_0| &&\\
& = 2\,|\cos(\delta_c)(\boldsymbol{i} \cdot (-\boldsymbol{i}/\norm{\boldsymbol{i}})) + c_0| &&\\
& = 2\,|-\norm{\boldsymbol{i}}\cos(\delta_c) + c_0|
\end{flalign*}
\noindent Similarly, we have:\\
$\norm{\boldsymbol{j} - \boldsymbol{m}(\boldsymbol{j},\mathcal{C})} = 2\,|\norm{\boldsymbol{j}}\cos(\delta_c) + c_0|$

\medskip 
\noindent If we assume that $\mathcal{C}$ lies between $\boldsymbol{i}$ and $\boldsymbol{j}$, then we must have $-\norm{\boldsymbol{i}} < c_0/\cos(\delta_c) < \norm{\boldsymbol{j}}$ and:\\
$\norm{\boldsymbol{i} - \boldsymbol{m}(\boldsymbol{i},\mathcal{C})} = 2\,(\norm{\boldsymbol{i}}\cos(\delta_c) - c_0)$\\
$\norm{\boldsymbol{j} - \boldsymbol{m}(\boldsymbol{j},\mathcal{C})} = 2\,(\norm{\boldsymbol{j}}\cos(\delta_c) + c_0)$

\medskip
\noindent By applying the law of cosines in the triangle $\boldsymbol{i}\,\boldsymbol{m}(\boldsymbol{i},\mathcal{C})\,\boldsymbol{j}$, we have:
\vspace{-0.1cm}
\begin{flalign*}
\norm{\boldsymbol{j} - \boldsymbol{m}(\boldsymbol{i},\mathcal{C})} & = \sqrt{\norm{\boldsymbol{i} - \boldsymbol{m}(\boldsymbol{i},\mathcal{C})}^2 + \norm{\boldsymbol{j} - \boldsymbol{i}}^2 - 2\,\norm{\boldsymbol{i} - \boldsymbol{m}(\boldsymbol{i},\mathcal{C})}\,\norm{\boldsymbol{j} - \boldsymbol{i}}\,\cos(\delta_c)} &&\\
& = \sqrt{4\,\left(\norm{\boldsymbol{i}}\cos(\delta_c) - c_0\right)^2 + 4\,\norm{\boldsymbol{i}}^2 - 8\,\left(\norm{\boldsymbol{i}}\cos(\delta_c) - c_0\right)\,\norm{\boldsymbol{i}}\,\cos(\delta_c)} &&\\
& = 2\,\sqrt{\norm{\boldsymbol{i}}^2\cos^2(\delta_c) + c_0^2 - 2\,\norm{\boldsymbol{i}}\cos(\delta_c)\,c_0 + \norm{\boldsymbol{i}}^2 - 2\,\norm{\boldsymbol{i}}^2\cos^2(\delta_c) + 2\,\norm{\boldsymbol{i}}\cos(\delta_c)\,c_0} &&\\
& = 2\,\sqrt{\norm{\boldsymbol{i}}^2\,(1 - \cos^2(\delta_c)) + c_0^2} &&\\
& = 2\,\sqrt{\norm{\boldsymbol{i}}^2\,\sin^2(\delta_c) + c_0^2}
\end{flalign*}
Similarly by applying the law of cosines in the triangle $\boldsymbol{j}\,\boldsymbol{m}(\boldsymbol{j},\mathcal{C})\,\boldsymbol{i}$, we have:\\
$\norm{\boldsymbol{i} - \boldsymbol{m}(\boldsymbol{j},\mathcal{C})} = 2\,\sqrt{\norm{\boldsymbol{j}}^2\,\sin^2(\delta_c) + c_0^2}$

\medskip
\noindent Finally by posing $r_c = c_0/\norm{\boldsymbol{i}} = c_0/\norm{\boldsymbol{j}} = 2\,c_0/\norm{\boldsymbol{j}-\boldsymbol{i}}$, we can write:\\[0.2cm]
$\displaystyle s(I,\mathcal{C}) = \arctan\left(\frac{\|\boldsymbol{j} - \boldsymbol{m}(\boldsymbol{i},\mathcal{C})\|}{\|\boldsymbol{i} - \boldsymbol{m}(\boldsymbol{i},\mathcal{C})\|}\right) = \arctan\left(\frac{\sqrt{\sin^2(\delta_c)+r_c^2}}{\cos(\delta_c)+r_c}\right)\\  
\displaystyle s(J,\mathcal{C}) = \arctan\left(\frac{\|\boldsymbol{i} - \boldsymbol{m}(\boldsymbol{j},\mathcal{C})\|}{\|\boldsymbol{j} - \boldsymbol{m}(\boldsymbol{j},\mathcal{C})\|}\right) = \arctan\left(\frac{\sqrt{\sin^2(\delta_c)+r_c^2}}{\cos(\delta_c)-r_c}\right)$

\subsubsection*{B \quad Expression of the sets of all classification scores through $\mathcal{C}$ and $\mathcal{C}_\theta$}

If we regard $S$ as a data matrix, then we can write:
\begin{flalign*}
d(S,\mathcal{C}) & = S \cdot \boldsymbol{c} + c_0 &&\\
& = S \cdot (\cos(\theta_c)\,\boldsymbol{z}^\perp_c + \sin(\theta_c)\,\boldsymbol{z}) + c_0 &&\\
& = \cos(\theta_c)\,(S \cdot \boldsymbol{z}^\perp_c) + \sin(\theta_c)\,(S \cdot \boldsymbol{z}) + c_0 &&\\
& = \cos(\theta_c)\,(S \cdot \boldsymbol{z}^\perp_c + c_0/\cos(\theta_c)) + \sin(\theta_c)\,(S \cdot \boldsymbol{z}) &&\\
& = (\cos(\theta_c),\; \sin(\theta_c)) \cdot S \cdot (\boldsymbol{z}^\perp_c + c_0/\cos(\theta_c),\; \boldsymbol{z})^\top &&\\
& = V \cdot P
\end{flalign*}
With $V = (\cos(\theta_c),\; \sin(\theta_c))$ and $P = S \cdot (\boldsymbol{z}^\perp_c + c_0/\cos(\theta_c),\; \boldsymbol{z})^\top$.

\bigskip
\noindent Similarly we have: $d(S,\mathcal{C}_\theta) = V_\theta \cdot P$

\smallskip
\noindent With $V_\theta = (\cos(\theta_c+\theta),\; \sin(\theta_c+\theta))$

\subsubsection*{C \quad Expression of $roc(\theta)$ when $P$ follows a bivariate normal distribution}

\underline{With covariance $\boldsymbol{\Sigma_1} = \diag(1,1)$:}

\begin{figure}[ht]
  \centering
  \includegraphics[width=0.66\textwidth]{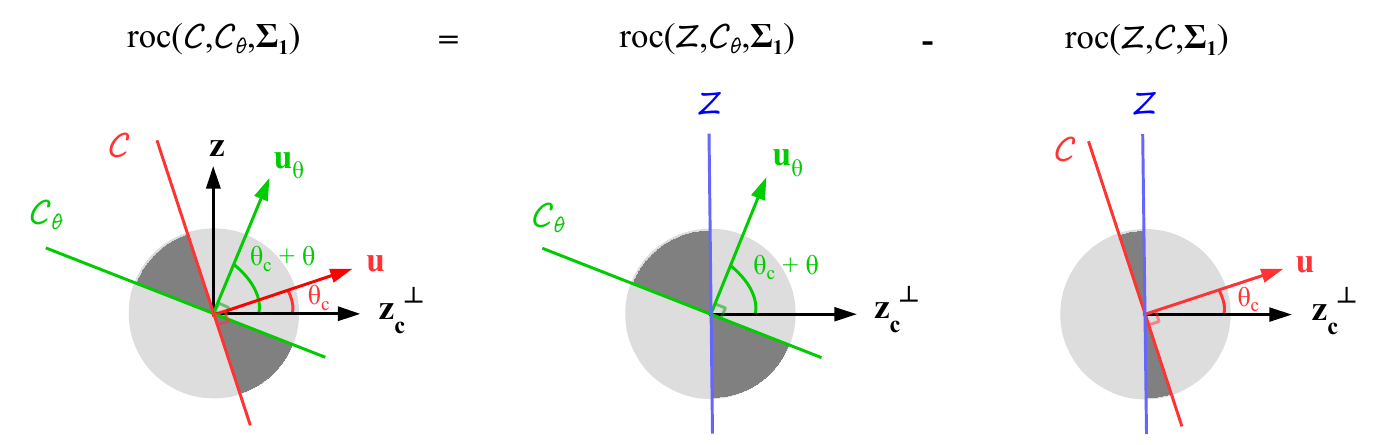}
  \label{roc2}
\end{figure}

\noindent $\displaystyle roc(\theta) = roc(\mathcal{C},\mathcal{C}_\theta,\boldsymbol{\Sigma_1}) = roc(\mathcal{Z},\mathcal{C}_\theta,\boldsymbol{\Sigma_1}) - roc(\mathcal{Z},\mathcal{C},\boldsymbol{\Sigma_1}) = \frac{\theta_c + \theta}{\pi} - \frac{\theta_c}{\pi} = \frac{\theta}{\pi}$

\bigskip
\noindent \underline{With covariance $\boldsymbol{\Sigma_2} = \diag(v_{\boldsymbol{z}}^\perp, \, v_{\boldsymbol{z}})$:}

\begin{figure}[ht]
  \centering
  \includegraphics[width=0.66\textwidth]{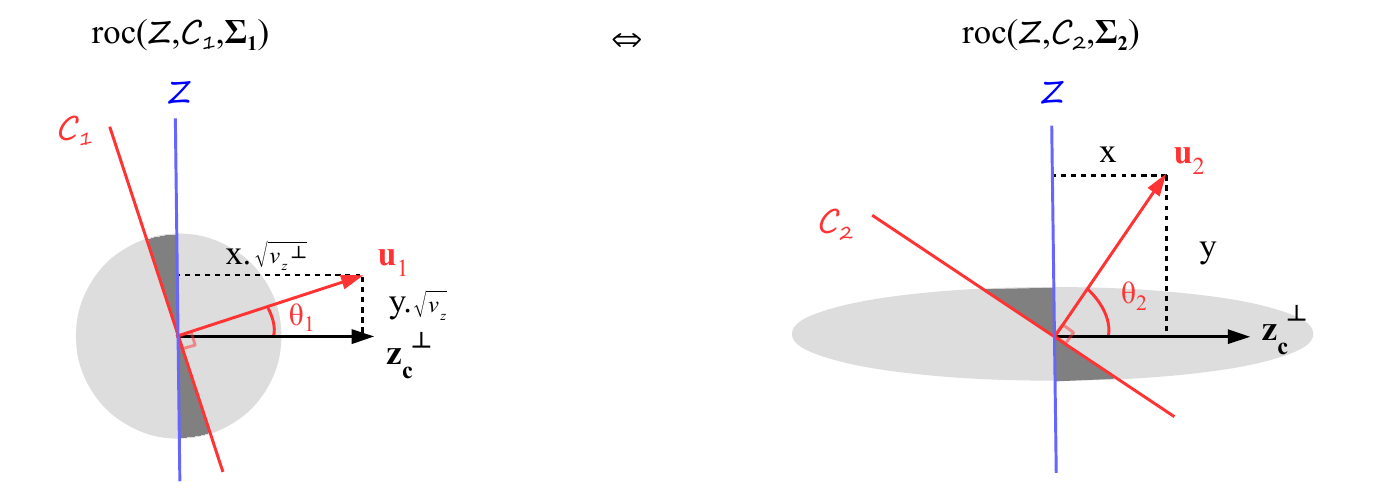}
  \label{}
\end{figure}

\noindent We have: 

\medskip
\noindent $\displaystyle roc(\mathcal{Z},\mathcal{C}_2,\boldsymbol{\Sigma_2}) = roc(\mathcal{Z},\mathcal{C}_1,\boldsymbol{\Sigma_1}) = \frac{\theta_1}{\pi}$

\bigskip
\noindent We also have: 

\medskip
\noindent $\displaystyle \tan(\theta_1) = \sqrt{\frac{v_{\boldsymbol{z}}}{v_{\boldsymbol{z}}^\perp}}\,\,\frac{y}{x} = \sqrt{\frac{v_{\boldsymbol{z}}}{v_{\boldsymbol{z}}^\perp}}\,\tan(\theta_2) \quad \Rightarrow \quad \theta_1 = \arctan\left(\sqrt{\frac{v_{\boldsymbol{z}}}{v_{\boldsymbol{z}}^\perp}}\,\tan(\theta_2)\right)$

\bigskip
\noindent Hence: 

\medskip
\noindent $\displaystyle roc(\mathcal{Z},\mathcal{C}_2,\boldsymbol{\Sigma_2}) = \frac{1}{\pi}\,\arctan\left(\sqrt{\frac{v_{\boldsymbol{z}}}{v_{\boldsymbol{z}}^\perp}}\,\tan(\theta_2)\right)$

\bigskip
\noindent And:
\begin{flalign*}
roc(\theta) & = roc(\mathcal{Z},\mathcal{C}_\theta,\boldsymbol{\Sigma_2}) - roc(\mathcal{Z},\mathcal{C},\boldsymbol{\Sigma_2}) &&\\
& = \frac{1}{\pi}\,\left[\arctan\left(\sqrt{\frac{v_{\boldsymbol{z}}}{v_{\boldsymbol{z}}^\perp}}\,\tan(\theta_c+\theta)\right) - \arctan\left(\sqrt{\frac{v_{\boldsymbol{z}}}{v_{\boldsymbol{z}}^\perp}}\,\tan(\theta_c)\right)\right] &&\\
& = \frac{1}{\pi}\left[\arctan\left(\sqrt{\frac{v_{\boldsymbol{z}}}{v_{\boldsymbol{z}}^\perp}}\tan(x)\right)\right]_{\theta_c}^{\theta_c+\theta}
\end{flalign*}

\newpage
\bibliographystyle{plainnat}
\bibliography{biblio}

\end{document}